\documentclass[runningheads]{llncs}
\usepackage{graphicx}

\usepackage{tikz}
\usepackage{comment}
\usepackage{amsmath,amssymb} 
\usepackage{color}

\usepackage{multicol}
\usepackage{multirow}
\usepackage{booktabs}

\usepackage[accsupp]{axessibility}  

\newlength\savewidth\newcommand\shline{\noalign{\global\savewidth\arrayrulewidth
  \global\arrayrulewidth 1pt}\hline\noalign{\global\arrayrulewidth\savewidth}}

\DeclareMathOperator{\avg}{avg}

\begin{document}
\pagestyle{headings}
\mainmatter
\def\ECCVSubNumber{4105}

\title{Polarimetric Pose Prediction
} 

\titlerunning{PPP-Net}
\author{Daoyi Gao$^{*}$ \and Yitong Li$^{*}$ \and Patrick Ruhkamp$^{*}$ \and Iuliia Skobleva$^{*}$ \and Magdalena Wysocki$^{*}$ \and HyunJun Jung \and Pengyuan Wang \and Arturo Guridi \and Benjamin Busam}
\authorrunning{D. Gao, Y. Li, P. Ruhkamp, I. Skobleva, M. Wysocki et al.}
\institute{\textit{$^{*}$ Equal contribution; Alphabetical order} \\
Technical University of Munich, Germany \\
\email{\tt\small \{d.gao,...,b.busam\}@tum.de}
}
\maketitle

\begin{abstract}
Light has many properties that vision sensors can passively measure.
Colour-band separated wavelength and intensity are arguably the most commonly used for monocular 6D object pose estimation.
This paper explores how complementary polarisation information, i.e. the orientation of light wave oscillations, influences the accuracy of pose predictions.
A hybrid model that leverages physical priors jointly with a data-driven learning strategy is designed and carefully tested on objects with different levels of photometric complexity.
Our design significantly improves the pose accuracy compared to state-of-the-art photometric approaches and enables object pose estimation for highly reflective and transparent objects.
A new multi-modal instance-level 6D object pose dataset with highly accurate pose annotations for multiple objects with varying photometric complexity is introduced as a benchmark. 
\end{abstract}

\section{Introduction}
"Fiat lux".\footnote{Latin for "let there be light".} Light has always fascinated humanity. It is not only the inherent centre of attention for many of the most significant scientific discoveries in the last century but also plays a crucial role in society and even sets the basis for religions. Typical light sensors in computer vision send or receive pulses and waves for which the wavelength and energy are measured to retrieve colour and intensity within a specified spectrum. However, intensity and wavelength are not the only properties of an electromagnetic (EM) wave. The oscillation direction of the EM-field relative to the light ray defines its polarisation. Most natural light sources such as the sun, a lamp or a candle emit unpolarised light, which means that the light wave oscillates in a multitude of directions. Light becomes perfectly or partially polarised when a wave is reflected off an object. Polarisation, therefore, carries information on surface structure, material and reflection angle, which can complement passively retrieved texture information from a scene~\cite{kalra2020deep}. These additional measurements are particularly interesting for photometrically challenging objects with metallic, reflective or transparent materials, which pose challenges to vision pipelines and effectively hamper their use for automation.
\begin{figure}[!t]
      \centering
      \includegraphics[width=1.0\columnwidth]{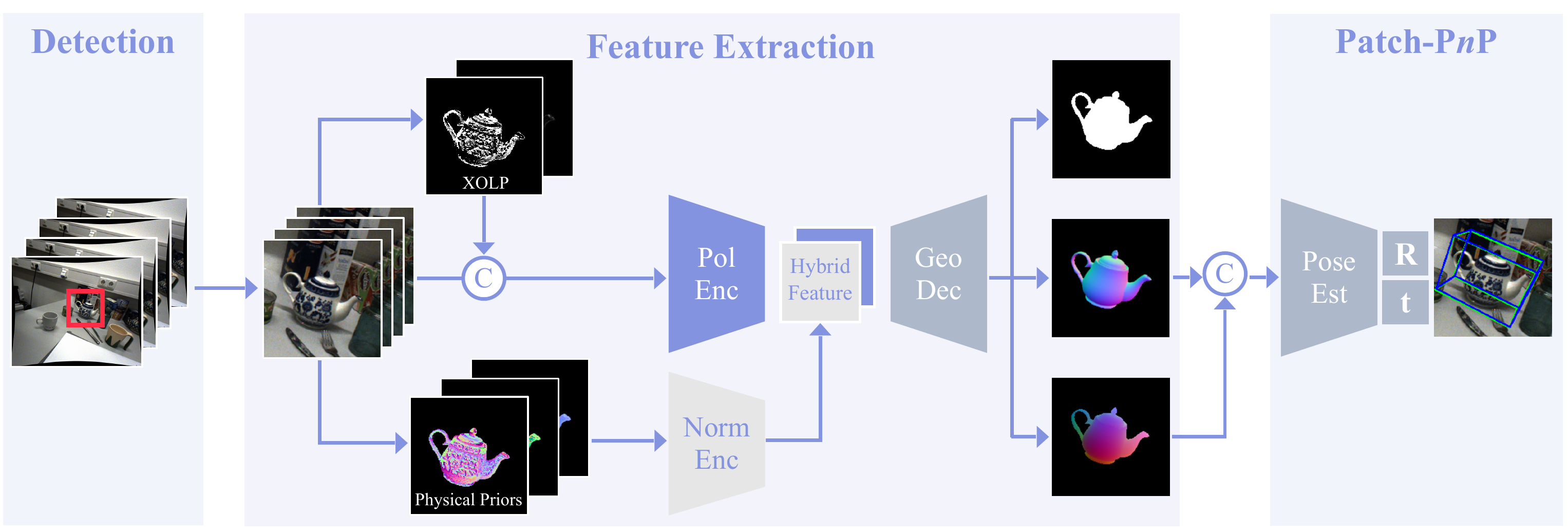}
      \caption{\textbf{PPP-Net. }
      Our \textbf{P}olarimetric \textbf{P}ose \textbf{P}rediction Pipeline utilises the RGBP images - a quadruple of four differently polarised RGB images - to compute AOLP/DOLP and polarised normal maps through our physical model. The polarised information and the physical cues are individually encoded and fused in our hybrid model. The decoder predicts object mask, normal map and NOCS, and finally the 6D object pose is predicted by Patch-P$n$P~\cite{wang2021gdr}.
      }
      \label{fig:teaser}
\end{figure}

While robust pipelines~\cite{hinterstoisser2012model,manhardt2019explaining,busam2020like,di2021so} have been designed for 6D pose estimation and texture-less~\cite{hodan2017t,drost2017introducing} objects have been successfully predicted, photometrically challenging objects with reflectance and partial transparency have just recently become the focus of research~\cite{liu2021stereobj}. These objects pose challenges to RGB-D sensing, and the field still lacks methods to cope with these problems. To address these limitations, we move beyond previous methods based on light intensity and exploit the polarisation properties of light as an additional prior for surface normals. This allows us to build a hybrid method combining a physical model with a data-driven learning approach to facilitate 6D pose estimation. We show that this not only supports pose estimation for photometrically challenging objects but also improves the pose accuracy for classical objects. To this end, our core contributions are \footnote{Dataset and code publicly available at: \url{https://daoyig.github.io/PPPNet/}.}:
\begin{enumerate}
    \item We propose \textbf{polarisation} as a new modality \textbf{for object pose estimation} and explore its advantages over previous modalities.
    \item We design a \textbf{hybrid pipeline} for instance-level 6D pose estimation that leverages polarisation cues through a \textbf{combination of physical model with learning}, which shows significant improvement for \textbf{photometrically challenging objects with high reflectance and translucency}.
    \item We construct the first \textbf{polarimetric instance-level 6D object pose estimation dataset} with highly accurate annotations.
\end{enumerate}
\section{Related Work}
\label{sec:related}
\subsection{Polarimetric Imaging}
\noindent\textbf{Polarisation for 2D. }
Polarisation cues provide valuable complementary information for various tasks in 2D computer vision that involve photometrically challenging objects. This has inspired a series of works on semantic~\cite{zhang2019exploration} and instance~\cite{kalra2020deep} segmentation for reflective and transparent objects. The absence of strong glare behind specific polarisation filters further helps to remove reflections from images~\cite{lei2020polarized}. While one polarisation camera can already provide significant improvements compared to photometric acquisition setups, multispectral polarimetric light fields~\cite{islam2021specular} boost the performance even more.

\noindent\textbf{Polarisation for 3D. }
Due to the inherent connection of polarisation with the object's surface, previous works on shape from polarisation (SfP) investigated the estimation of surface normals and depth from polarimetric data. However, intrinsic model ambiguities constrained setups in early works. Classical methods leverage an orthographic camera model and restrict the investigations to lab scenarios with controlled environment conditions~\cite{garcia2015surface,atkinson2006recovery,yu2017shape,smith2018height}. Yu et al.~\cite{yu2017shape} mathematically connect polarisation intensity with surface height and optimise for depth in a controlled scenario, while Atkinson et al.~\cite{atkinson2006recovery} recover surface orientation for fully diffuse surfaces. 
While these methods rely on monocular polarisation, more than one view can be combined with physical models for SfP~\cite{atkinson2005multi,cui2017polarimetric}, which can also be leveraged for self-supervision~\cite{CroMo}. Some works also explore the use of complementary photometric stereo~\cite{atkinson2017polarisation} and hybrid RGB+P approaches~\cite{zhu2019depth}, which complement each other and allow for metrically accurate depth estimates if the light direction is known. If an initial depth map exists, polarimetric cues can further refine the measurements~\cite{kadambi2017depth}. Furthermore, the polarimetric sensing model helps estimate the relative transformation of a moving polarisation sensor~\cite{cui2019polarimetric} assuming the scene is fully diffuse. Data-driven approaches can mitigate any assumptions on surface properties, light direction and object shapes. Ba et al.~\cite{ba2020deep} estimate surface normals by presenting a set of plausible cues to a neural network which uses these ambiguous cues for SfP. We take inspiration from this approach to complement our pose estimation pipeline with physical priors. In contrast to these works, we are interested in the object poses in an unconstrained setup without further assuming the reflection properties or lighting. The insights of previous works enable, for the first time, to design a pipeline which addresses pose prediction for photometrically challenging objects.

\subsection{6D Pose Prediction}
\noindent\textbf{Monocular RGB. }
Methods that predict 6D pose from a single image can be separated into three main categories: the ones that directly optimise for the pose, learn a pose embedding, or establish correspondences between the 3D model and the 2D image. Works that leverage pose parameterisation either directly regress the 6D pose~\cite{xiang2017posecnn,li2018deepim,manhardt2019explaining,labbe2020cosypose} or discretise the regression task and solve for classification~\cite{kehl2017ssd,busam2020like}. Networks trained this way directly predict pose parameters in the form of $SE\left(3\right)$ elements given the parameterisation used for training. Pose parameterisation can also be implicitly learnt~\cite{zhou2019continuity}. The second branch of methods~\cite{wohlhart2015learning,sundermeyer2018implicit,sundermeyer2020multi} utilises this to learn an implicit space to encode the pose from which the predictions can be decoded. The latest and also currently best-performing methods follow a two-stage approach. A network is used to predict 2D-3D correspondences between image and 3D model, which are used by a consecutive RANSAC/P$n$P pipeline that optimises the displacement robustly. Some methods in this field use sparse correspondences~\cite{rad2017bb8,peng2019pvnet,song2020hybridpose,hu2019segmentation}, while others establish dense 2D-3D pairs~\cite{zakharov2019dpod,park2019pix2pose,li2019cdpn,hodan2020epos}. While these methods typically learn the correspondences alone, some works learn the task end-to-end~\cite{hu2020single,wang2021gdr,di2021so}.
\noindent\textbf{RGB-D and Refinement. }
Since the task of monocular pose estimation from RGB is an inherently ill-posed problem, depth maps serve as a geometrical rescue. The spatial cue given by the depth map can be leveraged to establish point pairs for pose estimation~\cite{drost2010model} which can be further improved with RGB~\cite{birdal2015point}. In general, the pose can be recovered from depth or combined RGB-D, and most RGB-only methods (e.g.~\cite{sundermeyer2018implicit,li2019cdpn,park2019pix2pose,labbe2020cosypose}) benefit from a depth-driven refinement using ICP~\cite{besl1992method} or indirect multi-view cues~\cite{labbe2020cosypose}. The complementary information of RGB and depth has also inspired the seminal work DenseFusion~\cite{wang2019densefusion} in which deeply encoded features from both modalities are fused. FFB6D~\cite{He_2021_CVPR} further improves this through a tight coupling strategy with cross-modal information exchanges in multiple feature layers combined with a keypoint extraction~\cite{He_2020_CVPR} that leverages geometry and texture cues. These works, however, crucially depend on input quality, and depth-sensing suffers in photometrically challenging regions, where polarisation cues for depth could expedite the pose prediction. To the best of our knowledge, this has not been proposed yet.\\
\noindent\textbf{Photometric Challenges. }
The field of 6D pose estimation usually tests on well-established datasets with RGB-D input~\cite{hinterstoisser2012model,brachmann2014learning,xiang2017posecnn,kaskman2019homebreweddb}. Photometrically challenging objects such as texture-less and reflective industrial parts are also part of publically available datasets~\cite{hodan2017t,drost2017introducing}. While most of these datasets are carefully annotated for the pose, polarisation input is unavailable. Transparency is a further challenge addressed already in the pioneering work of Saxena et al.~\cite{saxena2008robotic} where the robotic grasp point of objects is determined from RGB stereo without a 3D model. Philipps et al.~\cite{phillips2016seeing} demonstrate how transparent objects with rotation symmetry can be reconstructed from two views using an edge detector and contour fitting. More recently, KeyPose~\cite{liu2020keypose} investigates instance and category level pose prediction from RGB stereo. Since their depth sensor fails on transparent objects, they leverage an opaque-transparent object pair to establish ground truth depth. ClearGrasp~\cite{sajjan2020clear} constitutes an RGB-D method that can be used on transparent objects. The recently available StereOBJ-1M dataset includes transparent, reflective and translucent objects with variations in illumination and symmetry using a binocular stereo RGB camera for pose estimation. However, none of these datasets comprised RGBP data.\\
\noindent To this end, the next natural step connects the shape cues from polarisation to recover object geometry in challenging environments. We further ask the question of how to do so by starting with a look into polarimetric image formation.

\section{Polarimetric Pose Prediction}
\begin{figure}[!b]
      \centering
      \includegraphics[width=0.5\linewidth]{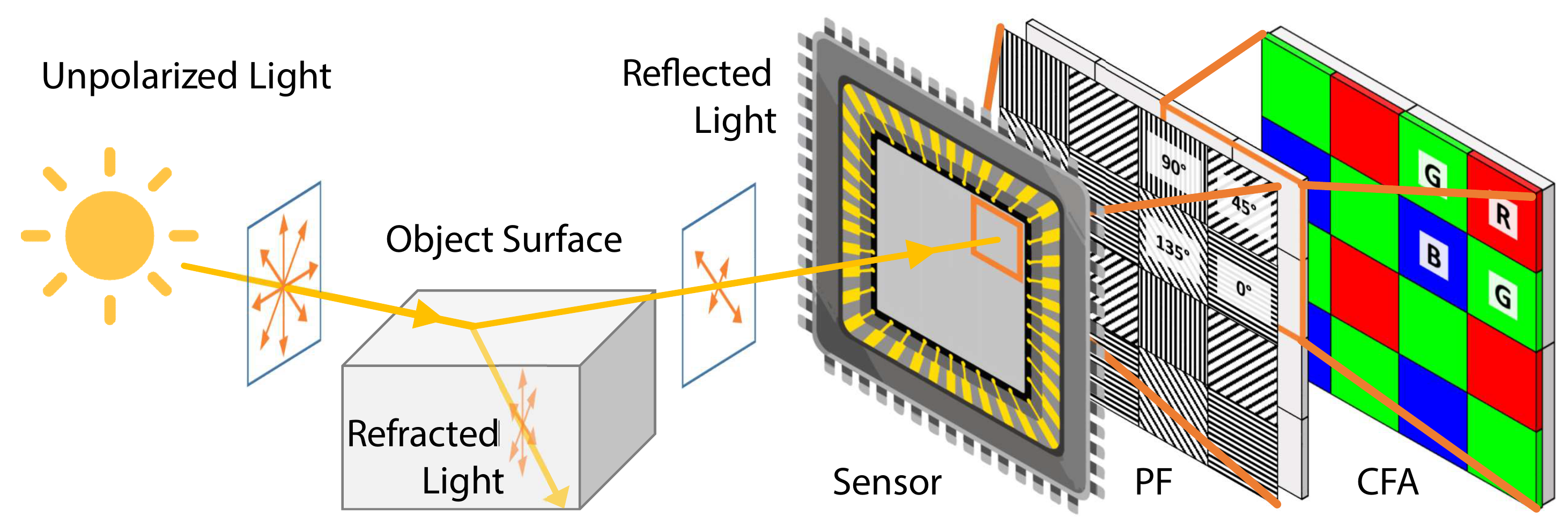}
      \caption{\textbf{Polarisation Camera. }
      When unpolarised light hits a surface, the refracted and reflected parts are partially polarised. A polarisation sensor captures the reflected light. In front of every pixel are four polarisation filters (PF) arranged at angles $0^{\circ}$, $45^{\circ}$, $90^{\circ}$, $135^{\circ}$. The colour filter array (CFA) separates light into different wavebands.
      }
      \label{fig:rgbp_sensor}
\end{figure}
In contrast to RGBP sensors (see Fig.~\ref{fig:rgbp_sensor}), RGB-D sensors enjoy wide use in the pose estimation field. Their cost-efficiency and tight integration in many devices present many possibilities in the vision field, but their design also comes with a few drawbacks.

\subsection{Photometric Challenges for RGB-D}
\label{sec:rgbd}
\begin{figure}[!t]
      \centering
      \includegraphics[width=0.5\columnwidth]{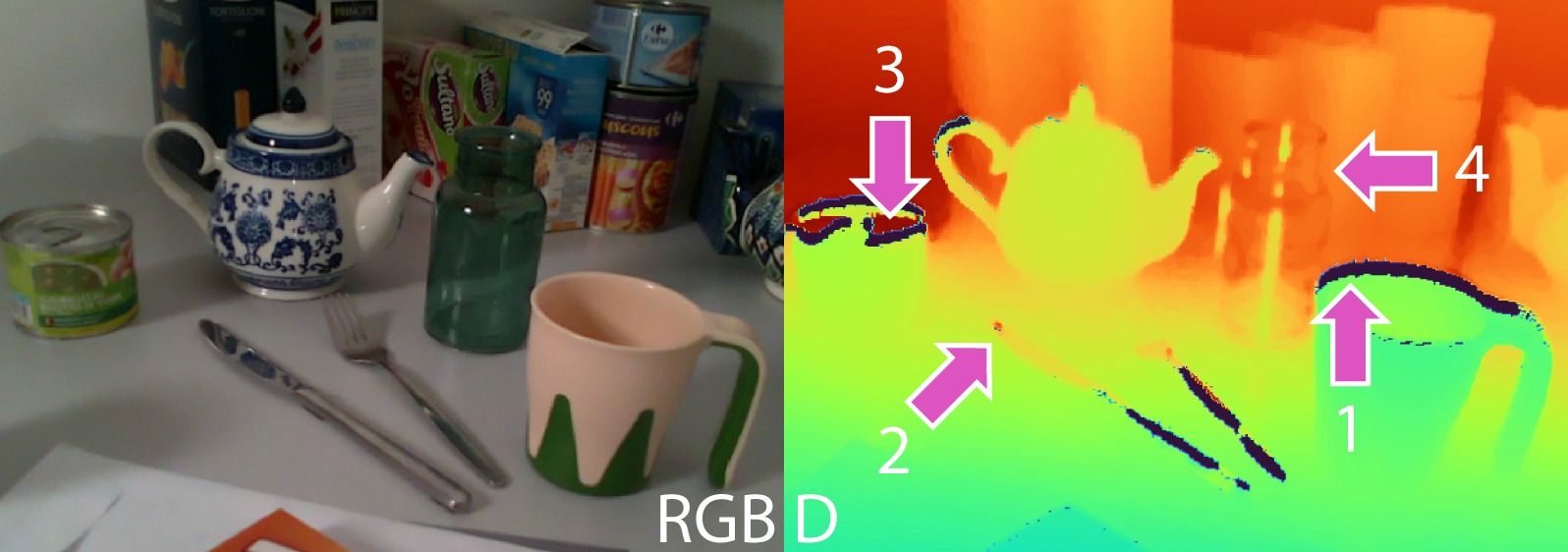}
      \caption{\textbf{Depth Artifacts. }
      A depth sensor miscalculates depth values for typical household objects. Reflective boundaries (1,3) invalidate pixels, while strong reflections (2,3) lead to incorrect values too far away. Semi-transparent objects (4) become partly invisible to the depth sensor, which measures the distance to the objects behind.
      }
      \label{fig:rgbd_issues}
\end{figure}

Commercial depth sensors typically use active illumination either by projecting a pattern (e.g. Intel RealSense D series) or using time-of-flight (ToF) measurements (e.g. Kinect v2 / Azure Kinect, Intel RealSense L series). While the former triangulates depth using stereo vision principles on projected or scene textures, the latter measures the roundtrip time of a light pulse that reflects from the scene. Since the measurement principle is photometric, both suffer on photometrically challenging surfaces where reflections artificially extend the roundtrip time of photons and translucent objects deteriorate the projected pattern to the extent that makes depth estimation infeasible. Fig.~\ref{fig:rgbd_issues} illustrates such an example for a set of common household objects. The semi-transparent vase becomes almost invisible to the used ToF sensor (RealSense L515). The reflections on both cutlery can lead to incorrect depth estimates significantly further than the correct value, while strong reflections at boundaries invalidate pixel distances.

\subsection{Surface Normals from Polarisation}
\label{sec:polarisation}
\begin{figure}[!b]
      \centering
      \includegraphics[width=0.5\columnwidth]{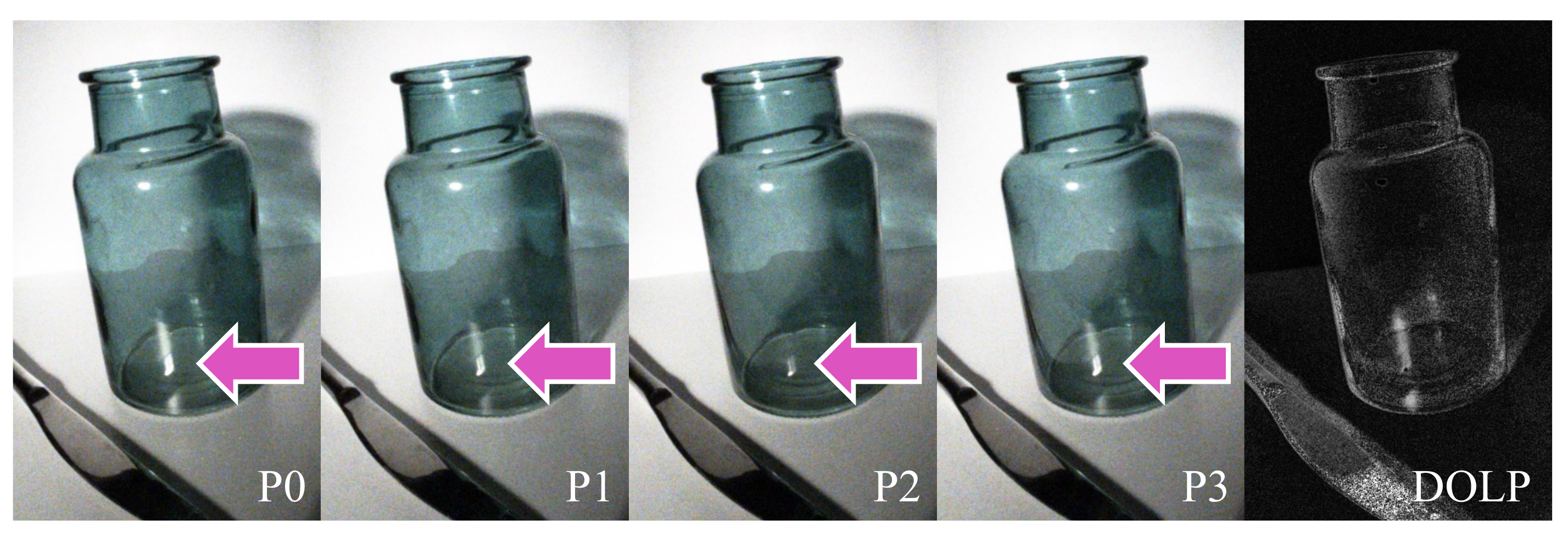}
      \caption{\textbf{DOLP. }
      Polarisation changes for the reflection of diffuse light on a translucent surface. Note the indicated differences in the polarimetric image quadruplet that directly relate to the surface normal. The degree of linear polarisation (DOLP) for the translucent and reflective surfaces is considerably higher than for the rest of the image.
      }
      \label{fig:pol_example}
\end{figure}

Before working with RGBP data, we introduce some physics behind polarimetric imaging.
Natural light and most artificially emitted light is unpolarised, meaning that the electromagnetic wave oscillates along all planes perpendicular to the direction of propagation of light~\cite{fliessbach2012elektrodynamik}. When unpolarised light passes through a linear polariser or is reflected at Brewster's angle from a surface, it becomes perfectly polarised. How fast light travels through the material and how much of it is reflected is determined by the \textit{refractive index}. It also determines Brewster's angle of that medium. When light is reflected at the same angle to the surface normal as the incident ray, we speak of \textit{specular reflection}. The remaining part penetrates the object as refracted light. As the light wave traverses through the medium, it becomes partially polarised. Following this, it escapes from the object and creates \textit{diffuse reflection}. For all real physical objects, the resulting reflection is a combination of specular and diffuse reflection, where the ratio largely depends on the refractive index and the angle of the incident light, as exemplified in Fig.~\ref{fig:pol_example}.

Light reaches the sensor with a specific intensity $I$ and wavelength $\lambda$. The sensor's colour filter array then separates the incoming light into RGB wavebands, as illustrated in Fig.~\ref{fig:rgbp_sensor}. The incoming light also has a degree of linear polarisation (DOLP) $\rho$ and a direction (angle) of polarisation (AOLP) $\phi$. The measured intensity behind a polariser with an angle $\varphi_{pol} \in \{0^\circ, 45^\circ, 90^\circ, 135^\circ\}$ depends on these parameters and the unpolarised intensity $I_{un}$~\cite{kalra2020deep}:
\begin{align}
    \label{eq:i_pol}
    I_{\varphi_{pol}} &= I_{un} \cdot \ (1+\rho \  \cos(2(\phi - \varphi_{pol}))).
\end{align}
We find $\varphi$ and $\rho$ from the over-determined system of linear equations in Eq.~\ref{eq:i_pol} using linear least squares. Depending on the surface properties, AOLP is calculated as:
\begin{align}
    \left\{
    \begin{array}{lll}
        \phi_{d} [\pi] &= \alpha \ &\text{for diffuse reflection}\\
        \phi_{s} [\pi] &= \alpha - \frac{\pi}{2} \ &\text{for specular reflection}
    \end{array}
    \right.
    ,
    \label{eqn:aolp}
\end{align}
where $[\pi]$ indicates the $\pi$-ambiguity and $\alpha$ is the azimuth angle of the surface normal $\textbf{n}$.
We can further relate the viewing angle $\theta \in [0, \pi/2]$ to the degree of polarisation by considering Fresnel coefficients thus DOLP is similarly given by~\cite{atkinson2006recovery}:
\begin{align}
    \left\{
    \begin{array}{l}
        \rho_{d} = \frac
            {(\eta-1/\eta)^{2}\sin^{2}(\theta)}
            {2+2\eta^{2}-(\eta+1/\eta)^{2}\sin^{2}(\theta)+4\cos(\theta)\sqrt{\eta^{2}-\sin^{2}(\theta)}}
            \\
            \\
        \rho_{s} =  \frac
            {2\sin^{2}(\theta)\cos(\theta)\sqrt{\eta^{2}-\sin^{2}(\theta)}}
            {\eta^{2}-\sin^{2}(\theta)- \eta^{2}\sin^{2}(\theta) +2\sin^{4}(\theta)}
    \end{array}
    \right.
    ,
    \label{eqn:dolp}
\end{align}
with the refractive index of the observed object material $\eta$. Solving Eq.~\ref{eqn:dolp} for $\theta$, we retrieve three solutions $\theta_d,\theta_{s1},\theta_{s2}$, one for the diffuse case and two for the specular case. For each of the cases, we can now find the 3D orientation of the surface by calculating the surface normals:
\begin{equation}
    \label{eq:normals}
    \mathbf{n} = \left(
    \cos{\alpha}\sin{\theta}, 
    \sin{\alpha}\sin{\theta},
    \cos{\theta}
    \right)^{\text{T}}
    .
\end{equation}
We use these plausible normals $\mathbf{n}_{d}, \mathbf{n}_{s1}, \mathbf{n}_{s2}$ as physical priors per pixel to guide our neural network to estimate the 6D object pose.

\subsection{Hybrid Polarimetric Pose Prediction Model}
\label{sec:method}
\label{sec:pol_pose_pipeline}
This section presents our \textbf{P}olarimetric \textbf{P}ose \textbf{P}rediction \textbf{Net}work, short \textbf{PPP-Net}. Given polarimetric images at four angles $I_{0}, I_{45}, I_{90}, I_{135}$, together with the calculated AOLP $\phi$, DOLP $\rho$, and normal maps $N_{d}, N_{s1}, N_{s2}$ as physical priors, we aim to utilise a network to learn a pose $\mathbf{P}=[\mathbf{R}|\mathbf{t}]$ transforms a target object from the object frame to the camera frame given a 3D CAD model of the object.

\noindent\textbf{Network Architecture. }
Our network architecture is depicted in Fig.~\ref{fig:teaser}. The first part of the network consists of two encoders with disjoint responsibilities. The first encodes joint polarisation information from native polarimetric images and the calculated AOLP/DOLP maps. The second one processes physical priors, i.e. the physical normals calculated from polarimetric images using the physical model. In both cases, we zoom in to a region of interest (ROI) of size $256 \times 256$ pixels. Then, the encoding is fused and passed to a decoder. The decoder receives the directly combined encoded information from both encoders enhanced by information from skip connections from different hierarchical levels of the encoders. Subsequently, it decodes an object mask, normal map, and a 3-channel dense correspondence map (NOCS) which creates a correspondence between each pixel and its normalised 3D coordinate. The predicted normal map and NOCS concatenated with corresponding 2D pixel coordinates are consecutively fed into a pose estimator as in~\cite{wang2021gdr}. The pose estimator comprises convolutional layers and fully connected layers and outputs the final estimated 3D rotation and translation.

\noindent\textbf{Pose Parametrisation. }
Inspired by recent works~\cite{zhou2019continuity,li2019cdpn,wang2021gdr} we parameterise the rotation as allocentric continuous 6D representation. Similarly, for translation we use a scale-invariant representation~\cite{li2019cdpn,wang2021gdr,di2021so}.\\
The continuous 6D representation $\mathbf{R}_\text{6d}$ for rotation comes from the first two columns of an original rotation matrix $\mathbf{R}$ \cite{zhou2019continuity} and we further turn it into allocentric representation~\cite{wang2021gdr,di2021so}. The allocentric representation is viewpoint-independent, and as such, it is favoured by our network, which only perceives the ROI of a target object. By reducing the scene to the zoomed-in ROI, we concentrate on the most relevant information in the image, i.e. our target object, which can facilitate improvement in the pose estimation. To overcome the limitations of a direct translation vector regression, we estimate the scale-invariant translation composed of relative differences between projected object centroids and the detected bounding box center location with respect to the bounding box size. The latter is given by $\delta_{x}, \delta_{y}$ and  the relative zoomed-in depth, $\delta_{z}$, where:
\begin{equation}
    \begin{cases}
    \delta_x =  (o_x - b_x) / b_w \\
    \delta_y =  (o_y - b_y) / b_h \\
    \delta_z =  t_z / r\\
    \end{cases}
    ,
    \label{eq:t_site}
\end{equation}
with $(o_x, o_y)$ and $(b_x, b_y)$ being the projected object centroids and bounding box center coordinates. The size of the bounding box $(b_w, b_h)$ is also used for calculating the zoomed-in ratio $r = s_{out} / s_{in}$ where $s_{in} = \max(b_w, b_h)$ and $s_{out}$ is the size of the output. Note that we can recover both the rotation matrix and translation vector with known camera intrinsics $K$~\cite{kundu20183d,li2019cdpn}.

\noindent\textbf{Object Normal Map. }
The surface normal map contains the surface orientation at each discrete pixel coordinate and thus encodes the shape of an object. Inspired by the previous works in SfP~\cite{ba2020deep}, we take a data-driven approach to retrieve the surface normal map. To better encode the geometric cue from the input physical priors apart from the polarisation cue, we do not concatenate the physical normals with the polarised images as Ba et al.~\cite{ba2020deep}, but encode them separately into two ResNet encoders. The decoder then learns to produce object shape encoded by the surface normal map. The estimated normals are L2-normalised to unit length. As shown in Tab.~\ref{tab:ablation}, with the given physical normals as shape prior, we can achieve high-quality normal map prediction, bringing a performance boost for the pose estimator. 

\noindent\textbf{Dense Correspondence Map. }
NOCS stores normalised 3D object coordinates given associated poses. This explicitly models correspondences between object 3D coordinates and projected 2D pixel locations. As shown by Wang et al.~\cite{wang2021gdr}, this representation helps a consecutive differentiable pose estimator achieve higher accuracy than RANSAC/P$n$P.

\subsection{Learning Objectives}
\label{sec:loss}
The overall objective is composed of both geometrical features learning and pose optimisation, as: $\mathcal{L} = \mathcal{L}_{pose} + \mathcal{L}_{geo}$, with:
\begin{gather}
    \label{eq:loss_pose_overall and loss_geo_overall}
    \mathcal{L}_{pose} = \mathcal{L}_{R} + \mathcal{L}_{center} + \mathcal{L}_{z} \\ 
    \mathcal{L}_{geo} = \mathcal{L}_{mask} + \mathcal{L}_{normals} + \mathcal{L}_{xyz}.
\end{gather}
Specifically, we employ separate loss terms for given ground truth rotation $\mathbf{R}$, $(\delta_x, \delta_y)$ and $\delta_z$ as:
\begin{equation}
    \begin{cases}
        \mathcal{L}_{R} &= \underset{\mathbf{x} \in \mathcal{M}}{\avg} \| {\mathbf{R}} \mathbf{x} - \hat{\mathbf{R}} \mathbf{x} \|_1 \\
        \mathcal{L}_{center} &= \| ({\delta}_x - \hat{\delta}_x,  {\delta}_y - \hat{\delta}_y) \|_1 \\
        \mathcal{L}_{z} &= \| {\delta}_z - \hat{\delta}_z \|_1
    \end{cases}
    ,
\label{eq:loss_pose_detail}
\end{equation}
where $\hat{\bullet}$ denotes prediction. For symmetrical objects, the rotation loss is calculated based on the smallest loss from all possible ground-truth rotations under symmetry.\\
To learn the intermediate geometrical features, we employ $L1$ losses for a mask and dense correspondences map learning and a cosine similarity loss for surface normal estimation:
\begin{equation}
    \begin{cases}
        \mathcal{L}_{mask} &= \| {\mathbf{M}}  - \hat{\mathbf{M}} \|_1 \\
        \mathcal{L}_{xyz} &= \mathbf{M} \odot \| \mathbf{M}_{xyz} - \hat{\mathbf{M}}_{xyz} \|_1 \\
        \mathcal{L}_{normal} &= 1- \langle\mathbf{n}, \hat{\mathbf{n}}\rangle
    \end{cases}
\label{eq:loss_geo_detail}
\end{equation}
\noindent where $\odot$ indicates the Hadamard product of element-wise multiplication, and $\langle\bullet,\bullet\rangle$ denotes the dot product.
\section{Polarimetric Data Acquisition}
We propose the first benchmark for 6D pose estimation through physical cues from polarimetric images for photometrically challenging objects. The objects in the dataset are chosen to cover a broad spectrum of photometric difficulties to yield scientifically meaningful insights: from matte to reflective and transparent. 

\begin{figure}[!b]
      \centering
      \includegraphics[width=0.5\columnwidth]{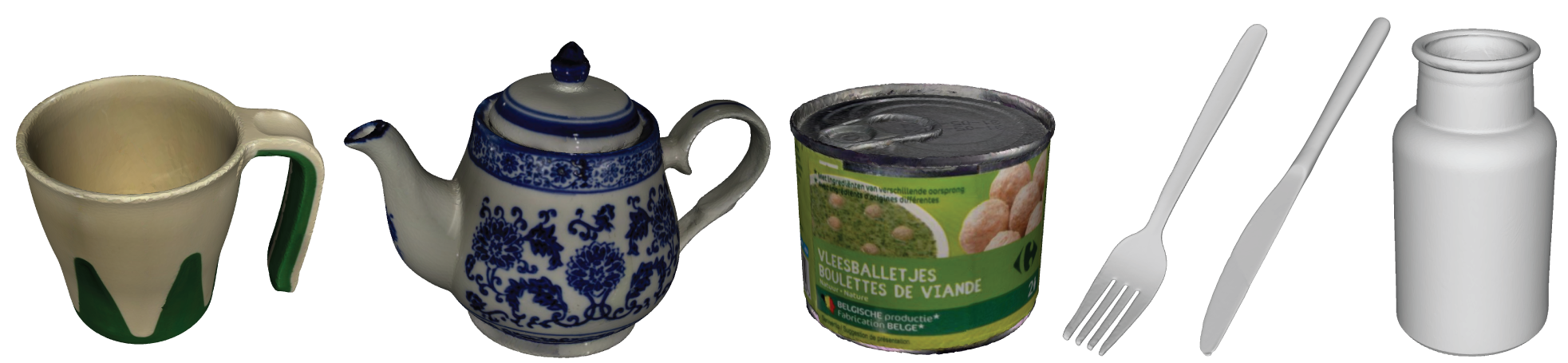}
      \caption{\textbf{3D Models. }
      Objects with increasing photometric complexity (left to right). Three objects have no texture due to reflection (cutlery) or transparency (bottle).
      }
      \label{fig:data_models}
\end{figure}

We follow the same data acquisition and annotation process as PhoCaL~\cite{PhoCal}, which is a category-level pose estimation dataset that comprises 60 household objects with high-quality 3D models scanned by a structured light 3D stereo scanner (EinScan-SP 3D Scanner, SHINING 3D Tech. Co., Ltd., Hangzhou, China). The scanning accuracy of the device is $\leq0.05$~mm which generates highly accurate models. We select the models \textit{cup}, \textit{teapot}, \textit{can}, \textit{fork}, \textit{knife}, \textit{bottle}, because of their increasing photometric complexity, as illustrated in Fig.~\ref{fig:data_models}. The last three models do not include texture due to their surface structure. Therefore we used a vanishing 3D scanning spray that made the surface temporarily opaque. To acquire RGB-D images, we use a direct Time-of-Flight (dToF) camera, Intel RealSense LiDAR Camera L515 (Intel, Santa Clara, California, USA), which captures RGB and Depth data at {640x480} pixel resolution.

RGBP data is acquired using the polarisation camera Phoenix 5.0 MP PHX 050S1-QC comprising a Sony IMX264MYR CMOS (Color) Polarsens sensor (LUCID Vision Labs, Inc., Richmond B.C, Canada) through a Universe Compact C-Mount 5MP 2/3'' 6mm f/2.0 lens (Universe, New York, USA) at {612x512} pixel resolution. Demosaicing is performed as part of the typical image signal processor (ISP) hardware pipeline, which is usually closed source (also for the commercial camera used here). Both cameras are mounted jointly to a KUKA iiwa (KUKA Roboter GmbH, Augsburg, Germany) 7~DoF robotic arm that guarantees a positional reproducibility of $\pm0.1$~mm. Intrinsic and extrinsic calibration is performed following a standard pinhole camera model~\cite{zhang2000flexible} with five distortion coefficients~\cite{heikkila1997four}. For pose annotation, we leverage a mechanical pose annotation method proposed in PhoCal~\cite{PhoCal} where a robot manipulator is used to tip the object of interest and extract a point cloud. This point cloud is consecutively aligned to the 3D model using ICP~\cite{besl1992method} to allow for highly accurate pose labels even for photometrically challenging objects. We plan a robot trajectory and use this setup to acquire four scenes with four different trajectories each and utilise a total of 8740 image sets for the dataset.

\section{Experimental Results}
\label{sec:experiment}
The motivation of our proposed pipeline is to show the advantage of leveraging pixelwise physical priors from polarised light (RGBP) for accurate 6D pose estimation of photometrically challenging objects - for which RGB-only and RGB-D methods often fail.
For this purpose, we train and test \textbf{PPP-Net} with different modalities first on two exemplary objects with very different levels of photometric complexity, i.e. a plastic \textit{cup}, and a photometrically very challenging, reflective and textureless stainless steel cutlery \textit{fork}. As detailed later, we find that polarimetric information yields significant performance gain for photometrically challenging objects.

\subsection{Experiments Setup}
\noindent\textbf{Implementation Details.}
We initially refine an off-the-shelf detector Mask RCNN~\cite{he2017mask} directly on the polarised images $I_{0}$ to provide useful object crops on our data (as is needed for the RGB-only benchmark and ours). We follow a similar training/testing split strategy as commonly used for the public datasets~\cite{brachmann2016uncertainty} and employ $\approx10\%$ of the RGBP images for training and $90\%$ for testing. We train our network end-to-end with Adam optimiser~\cite{kingma2014adam} for 200 epochs. The initial learning rate is set to 1e-4, halved every 50 epochs. As the depth sensor has a different field of view and is placed beneath the polarisation camera on a customised camera rig, the RGB-D benchmark split differs from the RGB training/testing split.

\noindent\textbf{Evaluation Metrics.}
To establish our proposed novel 6D pose estimation approach, we report the pose estimation accuracy per object as the commonly used average distance (ADD), and its equivalent for symmetrical objects (ADD-S)~\cite{hinterstoisser2012model} for different benchmarks. For the surface normal estimation, we calculate mean and median errors (in degrees) and the percentage of pixels where the estimated normals vary less than $11.25^\circ$, $22.5^\circ$ and $30^\circ$ from the ground truth.

\begin{table*}[!t]
\centering
\caption{\textbf{PPP-Net Modalities Evaluation. }
Different combinations of input and output modalities are used for training to study their influence on pose estimation accuracy ADD for objects with different photometric complexity. Where applicable, metrics for estimated normals are reported. Results for other objects in Supp. Mat.
}
\footnotesize
\resizebox{\textwidth}{!}{
\begin{tabular}{l|c|c|c|c|c|c|c|c|c|c|c|c}
\shline
\multirow{2}{*}{Object} & \multirow{2}{*}{\shortstack[c]{Photo.\\Chall.}} & \multicolumn{3}{c|}{Input Modalities} & \multicolumn{2}{c|}{Output Variants} & \multicolumn{5}{c|}{Normal Metrics} & Pose Metric \\
& \multicolumn{1}{c|}{} & \multicolumn{1}{c}{RGB} & \multicolumn{1}{c}{Polar RGB} & \multicolumn{1}{c|}{Physical N} & \multicolumn{1}{c}{Normals} & \multicolumn{1}{c|}{NOCS} & \multicolumn{1}{c}{mean$\downarrow$} & \multicolumn{1}{c}{med.$\downarrow$} & \multicolumn{1}{c|}{11.25$^\circ$$\uparrow$} & \multicolumn{1}{c|}{22.5$^\circ$$\uparrow$} & \multicolumn{1}{c|}{30$^\circ$$\uparrow$} & \multicolumn{1}{c}{ADD}\\
\hline
\multirow{4}{*}{Cup} & \multirow{4}{*}{} & \checkmark & & & & \checkmark & - & - & - & - & - & 91.1 \\
                                                                   & & & \checkmark & & & \checkmark & - & - & - & - & - & 91.3 \\
                                                                   & & & \checkmark & & \checkmark & \checkmark & 7.3 & 5.5 & 86.2 & 96.1 & 97.9 & 91.3 \\
                                                                   & & & \checkmark & \checkmark & \checkmark & \checkmark & \textbf{4.5} & \textbf{3.5} & \textbf{94.7} & \textbf{99.1} & \textbf{99.6} &  \textbf{97.2} \\
\hline
\multirow{4}{*}{Fork} & \multirow{4}{*}{\texttt{$\dagger\dagger$}} & \checkmark & & & & \checkmark & - & - & - & - & - & 85.4 \\
                                                                   & & & \checkmark & & & \checkmark & - & - & - & - & - & 86.1 \\
                                                                   & & & \checkmark & & \checkmark & \checkmark & 11.0 & 7.3 & 72.6 & 90.7 & 93.9 & 92.9\\
                                                                   & & & \checkmark & \checkmark & \checkmark & \checkmark & \textbf{6.5} & \textbf{4.3} & \textbf{87.6} & \textbf{95.9} & \textbf{97.6} &  \textbf{95.9} \\
\shline
\end{tabular}}
\label{tab:ablation}
\end{table*}

\begin{figure}[!b]
      \centering
      \includegraphics[width=0.8\columnwidth]{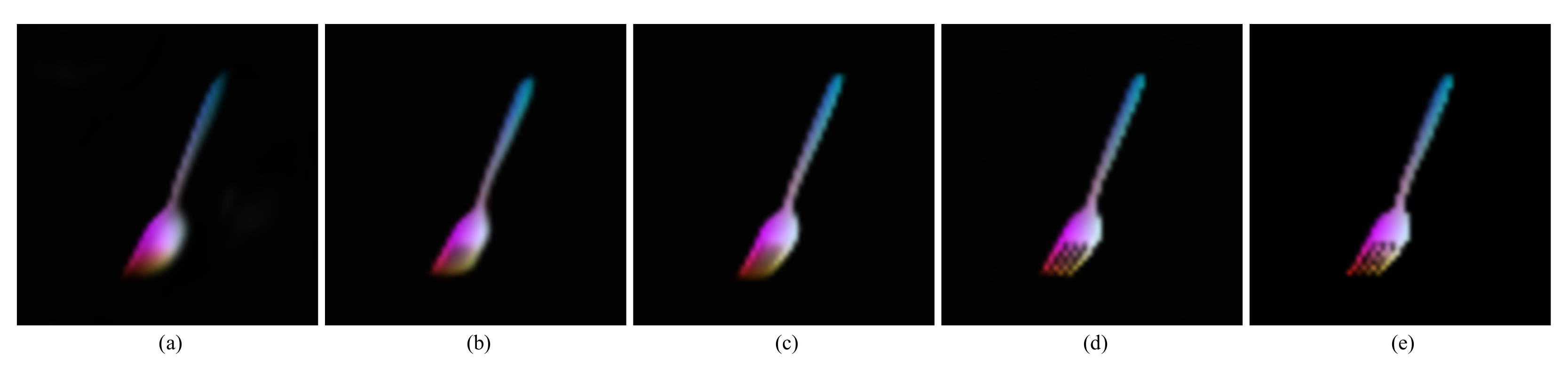}
      \caption{\textbf{Visualization of ablations on NOCS. }
      The quality of the geometrical representations improves when incorporating physical priors. The NOCS prediction from left to right follows the same order as the ablation experiments in Tab.\ref{tab:ablation}: (a) unpolarised RGB input, with NOCS output; (b) polarisation input, with NOCS output; (c) polarisation input, NOCS and normals output; (d) \textbf{ours:} full model with polarisation and physical priors input, NOCS and normals output; (e) GT NOCS.
      }
      \label{fig:ablation_fig}
\end{figure}

\subsection{PPP-Net Evaluation}
Here, we perform a series of experiments to study the influence of the input modality on the pose estimation accuracy (see Tab.~\ref{tab:ablation} for quantitative results, Fig.~\ref{fig:ablation_fig} for qualitative improvement of NOCS), where we specifically analyse the influence of polarimetric image information for the task of 6D pose estimation. To identify the direct influence of polarisation imaging for the task of accurate object pose estimation, we first establish an RGB-only baseline by neglecting our contributions of \textbf{PPP-Net}. To compute the unpolarised RGB image, we average over polarimetric images at complementary angles and use this as input for an RGB-only network. As shown in the first two rows in Tab.~\ref{tab:ablation} for each object (RGB against Polar RGB), the polarisation modality yields more considerable accuracy gains for the photometrically challenging object \textit{fork} as compared to \textit{cup}.

The accuracy of the pose estimator can be further improved when the network is guided to extract additional shape information of the object, which is implicitly encoded in the polarisation images (Tab.~\ref{tab:ablation}: 2nd to 3rd row). However, the quality of the output normals is limited. With the input of physically-induced normals from polarisation images, the network is provided with a plausible prior to encode shape information directly. Thus, it yields a much better normals prediction, significantly improving the pose performance (Tab.~\ref{tab:ablation}: 3rd to 4th row). The comparison of NOCS prediction shown in Fig.~\ref{fig:ablation_fig} reveals the fact that, given polarisation and direct shape cues, the network is guided to establish a more accurate and delicate geometrical representation, which is aligned with the quantitative improvement.

\begin{table*}[!t]
\centering
\caption{\textbf{Benchmark comparisons. } 
We compare our method against recent RGB-D (FFB6D~\cite{He_2021_CVPR}) and RGB-only (GDR-Net~\cite{wang2021gdr}) methods on a variety of objects with different level of photometric challenges ($\dagger$), and depth map quality (good: $+$ to low:$-$) which serves as input for FFB6D. RGB-D and RGB-only comparisons are trained and tested on different splits due to different field of view of depth camera (see Sec.~\ref{sec:experiment} for details). We report the Average Recall of ADD(-S).
}
\footnotesize
\resizebox{\textwidth}{!}{
\begin{tabular}{l|c|c|c|c|c|c|c|c|c||c|c}
\shline
\multirow{2}{*}{Object} & \multirow{2}{*}{\shortstack[c]{Photo.\\Chall.}} & \multicolumn{5}{c|}{\multirow{1}{*}{Properties}} & \multirow{2}{*}{\shortstack[c]{Depth\\Quality}} & \multicolumn{2}{c||}{RGB-D Split} & \multicolumn{2}{c}{RGB Split} \\
&& \multicolumn{1}{|c}{\multirow{1}{*}{Reflective}} & \multicolumn{1}{c}{\multirow{1}{*}{Metallic}} & \multicolumn{1}{c}{\multirow{1}{*}{Textureless}} & \multicolumn{1}{c}{\multirow{1}{*}{Transparent}} & \multicolumn{1}{c|}{\multirow{1}{*}{Symmetric}}
&& \multicolumn{1}{c}{\multirow{1}{*}{FFB6D}} & \multicolumn{1}{c||}{\multirow{1}{*}{\textbf{Ours}}} & \multicolumn{1}{c}{\multirow{1}{*}{GDR}} & \multicolumn{1}{c}{\multirow{1}{*}{\textbf{Ours}}}\\
\cline{1-10}\cline{11-12}
Cup & &&&&&& \texttt{(+)}& \textbf{99.4} & 98.1 & 96.7 & \textbf{97.2}\\
Teapot & \texttt{$\dagger$} & (*) && &  &  &\texttt{++} & 86.8 & \textbf{94.2} & 99.0 & \textbf{99.9}\\
\hline
Can & \texttt{$\dagger$} & * & * & & & & \texttt{-} & 80.4 & \textbf{99.7} & 96.5 & \textbf{98.4}\\
Fork & \texttt{$\dagger\dagger$} & * & * & * &  &  & \texttt{--} & 37.0 & \textbf{72.4} & 86.6 & \textbf{95.9}\\
Knife & \texttt{$\dagger\dagger$} & * & * & * & & &  \texttt{---} & 36.7 & \textbf{87.2} & 92.6 & \textbf{96.4}\\
\hline
Bottle & \texttt{$\dagger\dagger\dagger$} & * & & * & *  & * &  \texttt{None} & 61.5 & \textbf{93.6} & 94.4 & \textbf{97.5}\\
\hline
Mean & &&&&&&& 67.0 & \textbf{90.9} & 94.3 & \textbf{97.6} \\
\shline
\end{tabular}
}
\label{tab:baseline:rgbd}
\end{table*}

\subsection{Comparison with established Benchmarks}
The input modality experiments already demonstrate the robust capabilities of polarimetric imaging inputs for \textbf{PPP-Net} to successfully learn reliable 6D pose prediction with high accuracy for photometrically challenging objects. The depth map of an RGB-D sensor can also provide geometric information that can be utilised for the task of 6D object pose estimation. We compare our method against FFB6D~\cite{He_2021_CVPR}, which has a unique design that learns to combine appearance and depth information as well as local and global information from the two individual modalities. 

We train FFB6D on our data for each object individually and report the best ADD(-S) metric for all objects in Tab.~\ref{tab:baseline:rgbd}. The photometric challenge that each object constitutes is summarised in Tab.~\ref{tab:baseline:rgbd} and detailed by its properties (compare with Fig.~\ref{fig:data_models}). The objects are categorised into three classes based on the depth map quality of the depth sensor (also compare Fig.~\ref{fig:rgbd_issues}). We observe that objects with good depth maps and minor photometric challenges achieve high ADD values for FFB6D~\cite{He_2021_CVPR}. The increase in photometric complexity (and worse depth map quality) correlates with a decrease in ADD for challenging objects. The transparent \textit{Bottle} object is an exception to this pattern. The depth map is completely invalid (compare Fig.~\ref{fig:rgbd_issues}), but FFB6D still achieves high ADD. We hypothesise that the network successfully learns to ignore the depth map input from early training onward (see Sec.~\ref{sec:discussion} for details). \textbf{PPP-Net} achieves comparable results for easy objects and outperforms the strong benchmark for photometrically complex objects. Our method does not suffer from reduced ADD due to noisy or inaccurate depth maps but instead leverages the orthogonal surface information from RGBP data.

As \textbf{PPP-Net} profits vastly from physical priors from polarisation, we thoroughly investigate to which extent this additional information impacts the improvement of estimated poses, especially for photometrically challenging objects, by comparing the results also against the monocular RGB-only method GDR-Net~\cite{wang2021gdr}. We observe that while using polarimetric information slightly improves pose estimation accuracy for non-challenging objects, we can achieve superior performance for items with inconsistent photometric information due to reflection or transparency. In Tab.~\ref{tab:baseline:rgbd} the accuracy gain of \textbf{PPP-Net} against GDR-Net increases proportionally to the photometric complexity since our physical priors provide additional information about the geometry of an object.
\section{Discussion}
\label{sec:discussion}
\noindent\textbf{Limitations of current geometric methods. }
As mentioned earlier, we postulate that the RGB-D method ignores invalid depth data already in the early stages of training (e.g. for the transparent \textit{bottle)} and eventually learns to ignore noisy or corrupted depth information. To prove this assumption, we perform attacks on the input depth map for the FFB6D~\cite{He_2021_CVPR} encoder to analyse which parts of input modalities the network relies on when making a prediction. For this purpose, we add small Gaussian noise to the depth-related feature embedding in the bottleneck of the network and compare the ADD under this attack. We observe that the relative decrease is smaller for photometrically challenging objects as compared to objects with accurate depth maps ($27\%$ drop in ADD for \textit{knife} and $63\%$ for \textit{cup}). These findings suggest that the network indeed ignores the geometrical cues of inaccurate depth inputs.

\noindent\textbf{Benefits of Polarisation. }
We have shown that physical priors can significantly improve 6D pose estimation results for photometrically challenging objects. RGB-only methods do not incorporate any geometric information and therefore show worse results in scenarios with objects of little texture. Methods which try to leverage geometric priors from RGB-D~\cite{He_2021_CVPR} often cannot reliably recover the 6D pose of such objects, as the depth map is usually degenerated and corrupt. Our \textbf{PPP-Net}, as the first RGBP 6D object pose estimation method, successfully achieves learning accurate poses even for very challenging objects by extracting geometric information from physical priors. Qualitative results are shown in Figs.~\ref{fig:teaser} and~\ref{fig:qual}, and additionally in the supplementary material. Another benefit of using RGBP lies in the sensor itself: as the polarisation filter is directly integrated on the same sensor as the Bayer filter, both modalities are intrinsically calibrated, and the image can be acquired passively, paving the way to sensor integration on low-energy and mobile devices. RGB-D cameras, on the contrary, often require energy-expensive active illumination and extrinsic calibration, which prevents simple integration and introduces additional uncertainty to the final RGB-D image.

\noindent\textbf{Limitations. } 
Our physical model requires the knowledge of the refractive index to compute the physical priors reliably. To explore the potential of the physical model, unlike prior works ~\cite{smith2018height,ba2020deep} which fixed the refractive index to $\eta = 1.5$ for all experiments, we use physically plausible values according to the materials (we approximate the refractive index by using the look-up table provided by \url{https://refractiveindex.info/}). This means one needs to manually choose such parameters, which would limit the performance of the physical model when using objects with unknown composite materials. Moreover, substantial changes in texture also affect the reflection of light and thus DOLP calculation which, in turn, influences our physical priors. 
\section{Conclusion}
We have presented \textbf{PPP-Net}, the first learning-based 6D object pose estimation pipeline, which leverages geometric information from polarisation images through physical cues. Our method outperforms current state-of-the-art RGB-D and RGB methods for photometrically challenging objects and demonstrates at par performance for ordinary objects. Extensive ablations show the importance of complementary polarisation information for accurate pose estimation - specifically for objects without texture, i.e. reflective or transparent surfaces.

\begin{figure}[!hb]
      \centering
      \includegraphics[width=0.7\columnwidth]{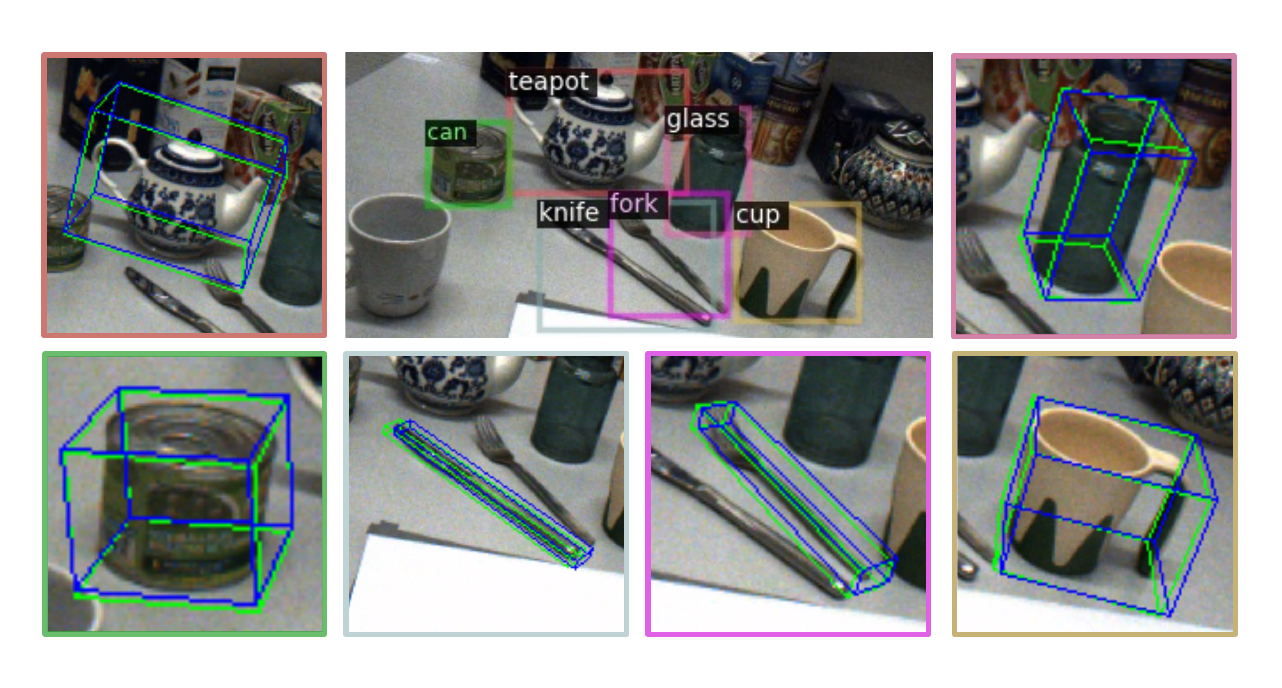}
      \caption{\textbf{Qualitative Results.} Input image with 2D detections are shown. Predicted and GT 6D poses are illustrated by \textit{blue} and \textit{green} bounding boxes, respectively.
      }
      \label{fig:qual}
\end{figure}

\newpage
\bibliographystyle{splncs04}
\bibliography{literature}

\begin{thebibliography}{10}
\providecommand{\url}[1]{\texttt{#1}}
\providecommand{\urlprefix}{URL }
\providecommand{\doi}[1]{https://doi.org/#1}

\bibitem{atkinson2017polarisation}
Atkinson, G.A.: Polarisation photometric stereo. Computer Vision and Image
  Understanding  \textbf{160},  158--167 (2017)

\bibitem{atkinson2005multi}
Atkinson, G.A., Hancock, E.R.: Multi-view surface reconstruction using
  polarization. In: IEEE International Conference on Computer Vision (ICCV).
  pp. 309--316 (2005)

\bibitem{atkinson2006recovery}
Atkinson, G.A., Hancock, E.R.: Recovery of surface orientation from diffuse
  polarization. Transactions on Image Processing  \textbf{15}(6),  1653--1664
  (2006)

\bibitem{ba2020deep}
Ba, Y., Gilbert, A., Wang, F., Yang, J., Chen, R., Wang, Y., Yan, L., Shi, B.,
  Kadambi, A.: Deep shape from polarization. In: European Conference on
  Computer Vision (ECCV). pp. 554--571 (2020)

\bibitem{besl1992method}
Besl, P.J., McKay, N.D.: Method for registration of 3d shapes. In: Sensor
  Fusion IV: Control Paradigms and Data Structures. vol.~1611, pp. 586--606.
  International Society for Optics and Photonics (1992)

\bibitem{birdal2015point}
Birdal, T., Ilic, S.: Point pair features based object detection and pose
  estimation revisited. In: IEEE International Conference on 3D Vision (3DV).
  pp. 527--535 (2015)

\bibitem{brachmann2014learning}
Brachmann, E., Krull, A., Michel, F., Gumhold, S., Shotton, J., Rother, C.:
  Learning 6d object pose estimation using 3d object coordinates. In: European
  Conference on Computer Vision (ECCV). pp. 536--551 (2014)

\bibitem{brachmann2016uncertainty}
Brachmann, E., Michel, F., Krull, A., Yang, M.Y., Gumhold, S., et~al.:
  Uncertainty-driven 6d pose estimation of objects and scenes from a single rgb
  image. In: Proceedings of the IEEE Conference on Computer Vision and Pattern
  Recognition (CVPR). pp. 3364--3372 (2016)

\bibitem{busam2020like}
Busam, B., Jung, H.J., Navab, N.: I like to move it: 6d pose estimation as an
  action decision process. arXiv preprint arXiv:2009.12678  (2020)

\bibitem{cui2017polarimetric}
Cui, Z., Gu, J., Shi, B., Tan, P., Kautz, J.: Polarimetric multi-view stereo.
  In: Proceedings of the IEEE Conference on Computer Vision and Pattern
  Recognition (CVPR). pp. 1558--1567 (2017)

\bibitem{cui2019polarimetric}
Cui, Z., Larsson, V., Pollefeys, M.: Polarimetric relative pose estimation. In:
  Proceedings of the IEEE/CVF International Conference on Computer Vision
  (ICCV). pp. 2671--2680 (2019)

\bibitem{di2021so}
Di, Y., Manhardt, F., Wang, G., Ji, X., Navab, N., Tombari, F.: So-pose:
  Exploiting self-occlusion for direct 6d pose estimation. In: Proceedings of
  the IEEE/CVF International Conference on Computer Vision (ICCV). pp.
  12396--12405 (2021)

\bibitem{drost2017introducing}
Drost, B., Ulrich, M., Bergmann, P., Hartinger, P., Steger, C.: Introducing
  mvtec itodd-a dataset for 3d object recognition in industry. In: Proceedings
  of the IEEE International Conference on Computer Vision (ICCV) Workshops. pp.
  2200--2208 (2017)

\bibitem{drost2010model}
Drost, B., Ulrich, M., Navab, N., Ilic, S.: Model globally, match locally:
  Efficient and robust 3d object recognition. In: IEEE Conference on Computer
  Vision and Pattern Recognition (CVPR). pp. 998--1005 (2010)

\bibitem{fliessbach2012elektrodynamik}
Flie{\ss}bach, T.: Elektrodynamik: Lehrbuch zur Theoretischen Physik II,
  vol.~2. Springer-Verlag (2012)

\bibitem{garcia2015surface}
Garcia, N.M., De~Erausquin, I., Edmiston, C., Gruev, V.: Surface normal
  reconstruction using circularly polarized light. Optics express
  \textbf{23}(11),  14391--14406 (2015)

\bibitem{he2017mask}
He, K., Gkioxari, G., Doll{\'a}r, P., Girshick, R.: Mask r-cnn. In: Proceedings
  of the IEEE International Conference on Computer Vision (ICCV). pp.
  2961--2969 (2017)

\bibitem{He_2021_CVPR}
He, Y., Huang, H., Fan, H., Chen, Q., Sun, J.: Ffb6d: A full flow bidirectional
  fusion network for 6d pose estimation. In: IEEE/CVF Conference on Computer
  Vision and Pattern Recognition (CVPR) (2021)

\bibitem{He_2020_CVPR}
He, Y., Sun, W., Huang, H., Liu, J., Fan, H., Sun, J.: Pvn3d: A deep point-wise
  3d keypoints voting network for 6dof pose estimation. In: IEEE/CVF Conference
  on Computer Vision and Pattern Recognition (CVPR) (2020)

\bibitem{heikkila1997four}
Heikkila, J., Silv{\'e}n, O.: A four-step camera calibration procedure with
  implicit image correction. In: Proceedings of IEEE Computer Society
  Conference on Computer Vision and Pattern Recognition (CVPR). pp. 1106--1112
  (1997)

\bibitem{hinterstoisser2012model}
Hinterstoisser, S., Lepetit, V., Ilic, S., Holzer, S., Bradski, G., Konolige,
  K., Navab, N.: Model based training, detection and pose estimation of
  texture-less 3d objects in heavily cluttered scenes. In: Asian Conference on
  Computer Vision (ACCV). pp. 548--562 (2012)

\bibitem{hodan2020epos}
Hodan, T., Barath, D., Matas, J.: Epos: Estimating 6d pose of objects with
  symmetries. In: Proceedings of the IEEE/CVF conference on computer vision and
  pattern recognition (CVPR). pp. 11703--11712 (2020)

\bibitem{hodan2017t}
Hodan, T., Haluza, P., Obdr{\v{z}}{\'a}lek, {\v{S}}., Matas, J., Lourakis, M.,
  Zabulis, X.: T-less: An rgb-d dataset for 6d pose estimation of texture-less
  objects. In: IEEE Winter Conference on Applications of Computer Vision
  (WACV). pp. 880--888 (2017)

\bibitem{hu2020single}
Hu, Y., Fua, P., Wang, W., Salzmann, M.: Single-stage 6d object pose
  estimation. In: Proceedings of the IEEE/CVF Conference on Computer Vision and
  Pattern Recognition (CVPR). pp. 2930--2939 (2020)

\bibitem{hu2019segmentation}
Hu, Y., Hugonot, J., Fua, P., Salzmann, M.: Segmentation-driven 6d object pose
  estimation. In: Proceedings of the IEEE/CVF Conference on Computer Vision and
  Pattern Recognition (CVPR). pp. 3385--3394 (2019)

\bibitem{islam2021specular}
Islam, M.N., Tahtali, M., Pickering, M.: Specular reflection detection and
  inpainting in transparent object through msplfi. Remote Sensing
  \textbf{13}(3), ~455 (2021)

\bibitem{kadambi2017depth}
Kadambi, A., Taamazyan, V., Shi, B., Raskar, R.: Depth sensing using
  geometrically constrained polarization normals. International Journal of
  Computer Vision (IJCV)  \textbf{125}(1-3),  34--51 (2017)

\bibitem{kalra2020deep}
Kalra, A., Taamazyan, V., Rao, S.K., Venkataraman, K., Raskar, R., Kadambi, A.:
  Deep polarization cues for transparent object segmentation. In: Proceedings
  of the IEEE/CVF Conference on Computer Vision and Pattern Recognition. pp.
  8602--8611 (2020)

\bibitem{kaskman2019homebreweddb}
Kaskman, R., Zakharov, S., Shugurov, I., Ilic, S.: Homebreweddb: Rgb-d dataset
  for 6d pose estimation of 3d objects. International Conference on Computer
  Vision (ICCV) Workshops  (2019)

\bibitem{kehl2017ssd}
Kehl, W., Manhardt, F., Tombari, F., Ilic, S., Navab, N.: Ssd-6d: Making
  rgb-based 3d detection and 6d pose estimation great again. In: Proceedings of
  the IEEE International Conference on Computer Vision (ICCV). pp. 1521--1529
  (2017)

\bibitem{kingma2014adam}
Kingma, D.P., Ba, J.: Adam: A method for stochastic optimization. arXiv
  preprint arXiv:1412.6980  (2014)

\bibitem{kundu20183d}
Kundu, A., Li, Y., Rehg, J.M.: 3d-rcnn: Instance-level 3d object reconstruction
  via render-and-compare. In: Proceedings of the IEEE Conference on Computer
  Vision and Pattern Recognition (CVPR). pp. 3559--3568 (2018)

\bibitem{labbe2020cosypose}
Labb{\'e}, Y., Carpentier, J., Aubry, M., Sivic, J.: Cosypose: Consistent
  multi-view multi-object 6d pose estimation. In: European Conference on
  Computer Vision (ECCV). pp. 574--591 (2020)

\bibitem{lei2020polarized}
Lei, C., Huang, X., Zhang, M., Yan, Q., Sun, W., Chen, Q.: Polarized reflection
  removal with perfect alignment in the wild. In: Proceedings of the IEEE/CVF
  Conference on Computer Vision and Pattern Recognition (CVPR). pp. 1750--1758
  (2020)

\bibitem{li2018deepim}
Li, Y., Wang, G., Ji, X., Xiang, Y., Fox, D.: Deepim: Deep iterative matching
  for 6d pose estimation. In: Proceedings of the European Conference on
  Computer Vision (ECCV). pp. 683--698 (2018)

\bibitem{li2019cdpn}
Li, Z., Wang, G., Ji, X.: Cdpn: Coordinates-based disentangled pose network for
  real-time rgb-based 6-dof object pose estimation. In: Proceedings of the
  IEEE/CVF International Conference on Computer Vision (ICCV). pp. 7678--7687
  (2019)

\bibitem{liu2021stereobj}
Liu, X., Iwase, S., Kitani, K.M.: Stereobj-1m: Large-scale stereo image dataset
  for 6d object pose estimation. In: Proceedings of the IEEE/CVF International
  Conference on Computer Vision (ICCV). pp. 10870--10879 (2021)

\bibitem{liu2020keypose}
Liu, X., Jonschkowski, R., Angelova, A., Konolige, K.: Keypose: Multi-view 3d
  labeling and keypoint estimation for transparent objects. In: Proceedings of
  the IEEE/CVF Conference on Computer Vision and Pattern Recognition (CVPR).
  pp. 11602--11610 (2020)

\bibitem{manhardt2019explaining}
Manhardt, F., Arroyo, D.M., Rupprecht, C., Busam, B., Birdal, T., Navab, N.,
  Tombari, F.: Explaining the ambiguity of object detection and 6d pose from
  visual data. In: Proceedings of the IEEE/CVF International Conference on
  Computer Vision (ICCV). pp. 6841--6850 (2019)

\bibitem{park2019pix2pose}
Park, K., Patten, T., Vincze, M.: Pix2pose: Pixel-wise coordinate regression of
  objects for 6d pose estimation. In: Proceedings of the IEEE/CVF International
  Conference on Computer Vision (ICCV). pp. 7668--7677 (2019)

\bibitem{peng2019pvnet}
Peng, S., Liu, Y., Huang, Q., Zhou, X., Bao, H.: Pvnet: Pixel-wise voting
  network for 6dof pose estimation. In: Proceedings of the IEEE/CVF Conference
  on Computer Vision and Pattern Recognition. pp. 4561--4570 (2019)

\bibitem{phillips2016seeing}
Phillips, C.J., Lecce, M., Daniilidis, K.: Seeing glassware: From edge
  detection to pose estimation and shape recovery. In: Robotics: Science and
  Systems. vol.~3 (2016)

\bibitem{rad2017bb8}
Rad, M., Lepetit, V.: Bb8: A scalable, accurate, robust to partial occlusion
  method for predicting the 3d poses of challenging objects without using
  depth. In: Proceedings of the IEEE International Conference on Computer
  Vision (ICCV). pp. 3828--3836 (2017)

\bibitem{sajjan2020clear}
Sajjan, S., Moore, M., Pan, M., Nagaraja, G., Lee, J., Zeng, A., Song, S.:
  Clear grasp: 3d shape estimation of transparent objects for manipulation. In:
  IEEE International Conference on Robotics and Automation (ICRA). pp.
  3634--3642 (2020)

\bibitem{saxena2008robotic}
Saxena, A., Driemeyer, J., Ng, A.Y.: Robotic grasping of novel objects using
  vision. The International Journal of Robotics Research  \textbf{27}(2),
  157--173 (2008)

\bibitem{smith2018height}
Smith, W.A., Ramamoorthi, R., Tozza, S.: Height-from-polarisation with unknown
  lighting or albedo. IEEE Transactions on Pattern Analysis and Machine
  Intelligence (T-PAMI)  \textbf{41}(12),  2875--2888 (2018)

\bibitem{song2020hybridpose}
Song, C., Song, J., Huang, Q.: Hybridpose: 6d object pose estimation under
  hybrid representations. In: Proceedings of the IEEE/CVF Conference on
  Computer Vision and Pattern Recognition (CVPR). pp. 431--440 (2020)

\bibitem{sundermeyer2020multi}
Sundermeyer, M., Durner, M., Puang, E.Y., Marton, Z.C., Vaskevicius, N., Arras,
  K.O., Triebel, R.: Multi-path learning for object pose estimation across
  domains. In: Proceedings of the IEEE/CVF Conference on Computer Vision and
  Pattern Recognition (CVPR). pp. 13916--13925 (2020)

\bibitem{sundermeyer2018implicit}
Sundermeyer, M., Marton, Z.C., Durner, M., Brucker, M., Triebel, R.: Implicit
  3d orientation learning for 6d object detection from rgb images. In:
  Proceedings of the European Conference on Computer Vision (ECCV). pp.
  699--715 (2018)

\bibitem{CroMo}
Verdie, Y., Song, J., Mas, B., Benjamin, B., Leonardis, A., , McDonagh, S.:
  Cromo: Cross-modal learning for monocular depth estimation. In: IEEE/CVF
  Conference on Computer Vision and Pattern Recognition (CVPR) (2022)

\bibitem{wang2019densefusion}
Wang, C., Xu, D., Zhu, Y., Mart{\'\i}n-Mart{\'\i}n, R., Lu, C., Fei-Fei, L.,
  Savarese, S.: Densefusion: 6d object pose estimation by iterative dense
  fusion. In: Proceedings of the IEEE/CVF Conference on Computer Vision and
  Pattern Recognition (CVPR). pp. 3343--3352 (2019)

\bibitem{wang2021gdr}
Wang, G., Manhardt, F., Tombari, F., Ji, X.: Gdr-net: Geometry-guided direct
  regression network for monocular 6d object pose estimation. In: Proceedings
  of the IEEE/CVF Conference on Computer Vision and Pattern Recognition (CVPR).
  pp. 16611--16621 (2021)

\bibitem{PhoCal}
Wang, P., Jung, H., Li, Y., Shen, S., Srikanth, R.P., Garattoni, L., Meier, S.,
  Navab, N., Busam, B.: Phocal: A multimodal dataset for category-level object
  pose estimation with photometrically challenging objects. In: IEEE/CVF
  Conference on Computer Vision and Pattern Recognition (CVPR) (2022)

\bibitem{wohlhart2015learning}
Wohlhart, P., Lepetit, V.: Learning descriptors for object recognition and 3d
  pose estimation. In: Proceedings of the IEEE Conference on Computer Vision
  and Pattern Recognition (CVPR. pp. 3109--3118 (2015)

\bibitem{xiang2017posecnn}
Xiang, Y., Schmidt, T., Narayanan, V., Fox, D.: Posecnn: A convolutional neural
  network for 6d object pose estimation in cluttered scenes. arXiv preprint
  arXiv:1711.00199  (2017)

\bibitem{yu2017shape}
Yu, Y., Zhu, D., Smith, W.A.: Shape-from-polarisation: A nonlinear least
  squares approach. In: Proceedings of the IEEE International Conference on
  Computer Vision (ICCV) Workshops. pp. 2969--2976 (2017)

\bibitem{zakharov2019dpod}
Zakharov, S., Shugurov, I., Ilic, S.: Dpod: 6d pose object detector and
  refiner. In: Proceedings of the IEEE/CVF International Conference on Computer
  Vision (ICCV). pp. 1941--1950 (2019)

\bibitem{zhang2019exploration}
Zhang, Y., Morel, O., Blanchon, M., Seulin, R., Rastgoo, M., Sidib{\'e}, D.:
  Exploration of deep learning-based multimodal fusion for semantic road scene
  segmentation. In: VISIGRAPP (5: VISAPP). pp. 336--343 (2019)

\bibitem{zhang2000flexible}
Zhang, Z.: A flexible new technique for camera calibration. Transactions on
  Pattern Analysis and Machine Intelligence (T-PAMI)  \textbf{22}(11),
  1330--1334 (2000)

\bibitem{zhou2019continuity}
Zhou, Y., Barnes, C., Lu, J., Yang, J., Li, H.: On the continuity of rotation
  representations in neural networks. In: Proceedings of the IEEE/CVF
  Conference on Computer Vision and Pattern Recognition (CVPR). pp. 5745--5753
  (2019)

\bibitem{zhu2019depth}
Zhu, D., Smith, W.A.: Depth from a polarisation + rgb stereo pair. In:
  Proceedings of the IEEE/CVF Conference on Computer Vision and Pattern
  Recognition. pp. 7586--7595 (2019)

\end{thebibliography}
\newpage

\section*{\centering\Large \textbf{PPP-Net -- Appendix}}

\setcounter{equation}{0}
\setcounter{figure}{0}
\setcounter{table}{0}
\setcounter{section}{0}

\renewcommand{\theequation}{A\arabic{equation}}
\renewcommand{\thefigure}{A\arabic{figure}}
\renewcommand{\thetable}{A\arabic{table}}
\renewcommand{\thesection}{A\arabic{section}}

\section{Physical Priors}
We use physical priors as inputs in our network to improve the estimated 6D pose of an object. These priors form relations between polarisation properties and azimuth and zenith angle of the surface normal, which serve as geometric cues orthogonal to color information. We calculate the physical priors under the assumption of either specular or diffuse reflection.
To recover the azimuth and zenith angle of the surface normal, we present the calculation for solving the unknowns of  Eq. \ref{eq:i_pol_appendix}.

A polarimetric camera registers intensity behind four linear polarisers with angles $0^\circ, 45^\circ, 90^\circ, 135^\circ$, which depends on unpolarised intensity $I_{un}$, degree of polarisation $\rho$, and angle of polarisation $\phi$:
\begin{align}
    \label{eq:i_pol_appendix}
    I_{\varphi_{pol}} &= I_{un} \cdot \ (1+\rho \  \cos(2(\phi - \varphi_{pol})))
    .
\end{align}
Eq. \ref{eq:i_pol_appendix} can be re-written as:
\begin{equation}
    \label{eq:i_pol_appendix_vector}
    I_{\varphi_{pol}} = 
    \underbrace{
    \begin{pmatrix}
        1\\
        \cos{2\varphi_{pol}}\\
        \sin{2\varphi_{pol}}
    \end{pmatrix}^T}_{\pmb{\beta^T}}
    \underbrace{
    \begin{pmatrix}
        I_{un}\\
        \rho\cos{2\phi}\\
        \rho\sin{2\phi}
    \end{pmatrix}}_{\pmb{x}
    }
    .
\end{equation}

For all angles $\varphi_{pol} \in \{0^\circ, 45^\circ, 90^\circ, 135^\circ \}$, we get a linear equation system for each pixel location with $\pmb{I_{\varphi_{pol}} \in {\rm I\!R}^{4\times1}}$, $\pmb{\beta \in {\rm I\!R}^{3\times4}}$ and $\pmb{x \in {\rm I\!R}^{3\times1}}$. After solving this over-determined linear equation system using least squares, we find unpolarised intensity, degree of polarisation and angle of polarisation:
\begin{equation}
    \begin{gathered}
    I_{un} = x_1 \\
    \rho = \sqrt{x_2^2 + x_3^2} \\
    \phi = \frac{1}{2}\arctan\frac{x_3}{x_2} 
    \end{gathered}
    .
\end{equation}

The azimuth angle can be found using Eq.2. Then, we can estimate the azimuth angle $\theta$ from Eq.3 by linear interpolation. Both models take in the same value for the refractive index $\eta$, since it is an intrinsic property of the material and it does not depend on the reflection model. The values used for our objects can be seen in Tab. \ref{tab:refindex}.

\begin{table}[!ht]
\centering
\caption{\textbf{Refractive Indices. }Refractive indices per object with certain material used for the physical model of \textbf{PPP-Net}.}
\resizebox{0.55\textwidth}{!}{
\begin{tabular}{l|c|c} 
    \shline
    Object & Material & Refractive Index \\ 
    \hline
    Teapot & ceramic & 1.54 \\ 
    Can & aluminium composite & 1.35 \\ 
    Fork & stainless steel & 2.75 \\ 
    Knife & stainless steel & 2.75 \\ 
    Bottle & glass & 1.52 \\ 
    Cup & plastics & 1.50 \\ 
    \shline
\end{tabular}
}
\label{tab:refindex}
\end{table}

\section{Additional Experiments and Ablation Studies} 
\noindent\textbf{Runtime Analysis.} On a desktop PC with an Intel i7 4.20GHz CPU and an NVIDIA 2080 GPU, given a $512 \times 612$ pixel image, our network takes ca.~64 ms for a single object, including 40 ms for detection, and 13 ms to calculate the physical priors with our non-optimized implementation.

\subsection{Ablations on Modalities}
\noindent\textbf{Ablations on Input Modalities. }
Tab.~\ref{tab:ablation_full} is an extension to Tab.1 in the main paper and summarises the quantitative evaluation for different modalities for \textbf{PPP-Net} for all objects under consideration in the dataset.

\begin{table*}[!ht]
\centering
\caption{\textbf{PPP-Net Input Modalities Evaluation. }
Different combinations of input and output modalities are used for training to study their influence on pose estimation accuracy ADD(-S) for objects with different photometric complexity. Where applicable, metrics for estimated normals are reported as well.
}

\resizebox{\textwidth}{!}{
\begin{tabular}{l|c|c|c|c|c|c|c|c|c|c|c|c}
\shline
\multirow{2}{*}{Object} & \multirow{2}{*}{\shortstack[c]{Photo.\\Chall.}} & \multicolumn{3}{c|}{Input Modalities} & \multicolumn{2}{c|}{Output Variants} & \multicolumn{5}{c|}{Normal Metrics} & Pose Metric \\
& \multicolumn{1}{c|}{} & \multicolumn{1}{c}{RGB} & \multicolumn{1}{c}{Polar RGB} & \multicolumn{1}{c|}{Physical N} & \multicolumn{1}{c}{Normals} & \multicolumn{1}{c|}{NOCS} & \multicolumn{1}{c}{mean$\downarrow$} & \multicolumn{1}{c}{med.$\downarrow$} & \multicolumn{1}{c|}{11.25$^\circ$$\uparrow$} & \multicolumn{1}{c|}{22.5$^\circ$$\uparrow$} & \multicolumn{1}{c|}{30$^\circ$$\uparrow$} & \multicolumn{1}{c}{ADD(-S)}\\
\hline
\multirow{4}{*}{Cup} & \multirow{4}{*}{} & \checkmark & & & & \checkmark & - & - & - & - & - & 91.1 \\
                                                                   & & & \checkmark & & & \checkmark & - & - & - & - & - & 91.3 \\
                                                                   & & & \checkmark & & \checkmark & \checkmark & 7.3 & 5.5 & 86.2 & 96.1 & 97.9 & 91.3 \\
                                                                   & & & \checkmark & \checkmark & \checkmark & \checkmark & \textbf{4.5} & \textbf{3.5} & \textbf{94.7} & \textbf{99.1} & \textbf{99.6} &  \textbf{97.2} \\
\hline
\multirow{4}{*}{Teapot} & \multirow{4}{*}{\texttt{$\dagger$}} & \checkmark & & & & \checkmark & - & - & - & - & - & 97.8 \\
                                                                   & & & \checkmark & & & \checkmark & - & - & - & - & - & 99.5 \\
                                                                   & & & \checkmark & & \checkmark & \checkmark & 7.9 & 5.4 & 82.5 & 94.5 & 97.1 & 99.2 \\
                                                                   & & & \checkmark & \checkmark & \checkmark & \checkmark & \textbf{5.3} & \textbf{4.0} & \textbf{91.6} & \textbf{98.7} & \textbf{99.5} &  \textbf{99.9} \\
                                                                   \hline
\multirow{4}{*}{Can} & \multirow{4}{*}{\texttt{$\dagger$}} & \checkmark & & & & \checkmark & - & - & - & - & - & 91.8 \\
                                                                   & & & \checkmark & & & \checkmark & - & - & - & - & - & 93.2 \\
                                                                   & & & \checkmark & & \checkmark & \checkmark & \textbf{5.7} & \textbf{3.9} & \textbf{90.0} & 97.0 & 98.6 & 96.7 \\
                                                                   & & & \checkmark & \checkmark & \checkmark & \checkmark & 6.0 & 4.5 & 89.0 & \textbf{97.3} & \textbf{98.9} &  \textbf{98.4} \\
\hline
\multirow{4}{*}{Fork} & \multirow{4}{*}{\texttt{$\dagger\dagger$}} & \checkmark & & & & \checkmark & - & - & - & - & - & 85.4 \\
                                                                   & & & \checkmark & & & \checkmark & - & - & - & - & - & 86.1 \\
                                                                   & & & \checkmark & & \checkmark & \checkmark & 11.0 & 7.3 & 72.6 & 90.7 & 93.9 & 92.9\\
                                                                   & & & \checkmark & \checkmark & \checkmark & \checkmark & \textbf{6.5} & \textbf{4.3} & \textbf{87.6} & \textbf{95.9} & \textbf{97.6} &  \textbf{95.9} \\
\hline
\multirow{4}{*}{Knife} & \multirow{4}{*}{\texttt{$\dagger\dagger$}} & \checkmark & & & & \checkmark & - & - & - & - & - & 84.1 \\
                                                                   & & & \checkmark & & & \checkmark & - & - & - & - & - & 88.0 \\
                                                                   & & & \checkmark & & \checkmark & \checkmark & 12.2 & 8.0 & 68.7 & 88.5 & 92.4 & 89.4\\
                                                                   & & & \checkmark & \checkmark & \checkmark & \checkmark & \textbf{6.8} & \textbf{5.4} & \textbf{88.2} & \textbf{97.3} & \textbf{98.6} &  \textbf{96.4} \\
                                                                   \hline
\multirow{4}{*}{Bottle} & \multirow{4}{*}{\texttt{$\dagger\dagger\dagger$}} & \checkmark & & & & \checkmark & - & - & - & - & - & 90.5 \\
                                                                  & & & \checkmark & & & \checkmark & - & - & - & - & - & 93.5 \\
                                                                   & & & \checkmark & & \checkmark & \checkmark & 5.6 & 4.7 & \textbf{92.9} & \textbf{99.0} & \textbf{99.6} & 94.7\\
                                                                   & & & \checkmark & \checkmark & \checkmark & \checkmark & \textbf{5.4} & \textbf{4.5} & 92.1 & \textbf{99.0} & \textbf{99.6} &  \textbf{97.5} \\
\shline
\end{tabular}}
\label{tab:ablation_full}
\end{table*}

\noindent\textbf{Ablations on Output Modalities. }
6D pose estimation mainly depends on accurate correspondences prediction by NOCS regression as reported in the ablation in Tab.~\ref{tab:ablation_nocs}.
The ADD drops significantly for the model without (w/o) NOCS output before Patch-PnP, i.e. only shape information is utilised for pose prediction.
Still, as proven by the ablations in the paper, the auxiliary explicit prediction of object-centric shape information as normals map benefits 6D pose estimation as the network is more strongly guided towards extracting physical shape priors from the input.

\begin{table}[!h]
\caption{\textbf{PPP-Net Output Ablation. }With and without NOCS output. 
}
\centering

\resizebox{0.35\columnwidth}{!}{
\begin{tabular}{l|c|c}
\shline
\multirow{1}{*}{Object} & \multicolumn{2}{c}{Pose Metric (ADD)} \\
\hline
\multirow{1}{*}{Teapot} & w/ \textbf{99.9} & w/o 72.7 \\
\hline
\multirow{1}{*}{Fork} & w/ \textbf{95.9} & w/o 79.3 \\
\shline
\end{tabular}}
\label{tab:ablation_nocs}
\end{table}

\subsection{Ablations on Network Architecture}
Tab.~\ref{tab:fusion_type} indicates naively concatenating geometric priors and RGBP images for direct input to the network (as in [5]) results in inferior normal prediction quality, and also leads to less improvement on pose estimation results (compare \texttt{concat} against \texttt{ours} in Tab.~\ref{tab:fusion_type}). This holds true for all objects, whereas photometrically more challenging objects show a larger relative improvement. These results confirm the importance of our design choices of \textbf{PPP-Net} to employ a dedicated encoder for the physics-based derived geometric priors, and its positive effect on 6D object pose estimation results.
We thus propose a careful integration design of such physical priors into established principles of 6D object pose estimation within our novel hybrid encoder. 
We deliberately choose a simple general architecture for PPP-Net for best comparison and evaluation against SOTA, and to show that even such simplistic encoders can achieve significant accuracy for 6D pose prediction with the physical priors from polarisation as inputs.

\begin{table}[!h]
\centering
\caption{\textbf{Fusion Ablation. }Naive concatenation against our proposed fusion strategy of RGB and physical priors in \textbf{PPP-Net}.}

\resizebox{1.0\columnwidth}{!}{
\begin{tabular}{l|c|c|c|c|c|c|c|c|c|c|c}
\hline
\multirow{2}{*}{Object} & \multirow{2}{*}{\shortstack[c]{Fusion}} & \multicolumn{2}{c|}{Input Modalities} & \multicolumn{2}{c|}{Output Variants} & \multicolumn{5}{c|}{Normal Metrics} & Pose Metric \\
& \multicolumn{1}{c|}{} & 
\multicolumn{1}{c|}{Polar RGB} & 
\multicolumn{1}{c|}{Physical N} & 
\multicolumn{1}{c|}{Normals} & \multicolumn{1}{c|}{NOCS} & \multicolumn{1}{c|}{mean$\downarrow$} & \multicolumn{1}{c|}{med.$\downarrow$} & \multicolumn{1}{c|}{11.25$^\circ$ $\uparrow$} & \multicolumn{1}{c|}{22.5$^\circ$$\uparrow$} & \multicolumn{1}{c|}{30$^\circ$$\uparrow$} & \multicolumn{1}{c}{ADD}\\
\hline
Cup & \texttt{concat} & \checkmark & \checkmark & \checkmark & \checkmark & 6.0 & 4.9 & 91.1
 & 98.1 & 99.1 & 93.6 \\
Cup & \texttt{ours} & \checkmark & \checkmark & \checkmark & \checkmark & \textbf{4.5} & \textbf{3.5} & 
\textbf{94.7} & \textbf{99.1} & \textbf{99.6} & \textbf{97.2}\\
\hline
Teapot & \texttt{concat} & \checkmark & \checkmark & \checkmark & \checkmark & 7.4 & 5.7 & 83.4
 & 96.3 & 98.4 & 97.3\\
Teapot & \texttt{ours} & \checkmark & \checkmark & \checkmark & \checkmark & \textbf{5.3} & \textbf{4.0} & 
\textbf{91.6} & \textbf{98.7} & \textbf{99.5} & \textbf{99.9}\\
\hline
Can & \texttt{concat} & \checkmark & \checkmark & \checkmark & \checkmark & 8.5 & 6.4 & 81.8
 & 95.1 & 97.5 & 92.2\\
Can & \texttt{ours} & \checkmark & \checkmark & \checkmark & \checkmark & \textbf{6.0} & \textbf{4.5} & 
\textbf{89.0} & \textbf{97.3} & \textbf{98.9} & \textbf{98.4}\\
\hline
Fork & \texttt{concat} & \checkmark & \checkmark & \checkmark & \checkmark &  10.7&  7.8& 
 70.0&  91.8&  95.0& 87.6\\
Fork & \texttt{ours} & \checkmark & \checkmark & \checkmark & \checkmark & \textbf{6.5} & \textbf{4.3} & 
\textbf{87.6} & \textbf{95.9} & \textbf{97.6} & \textbf{95.9}\\
\hline
Knife & \texttt{concat} & \checkmark & \checkmark & \checkmark & \checkmark & 10.8 & 8.5 & 
67.1 & {92.8} & {96.2} & 86.1\\
Knife & \texttt{ours} & \checkmark & \checkmark & \checkmark & \checkmark & \textbf{6.8} & \textbf{5.4} & 
\textbf{88.2} & \textbf{97.3} & \textbf{98.6} & \textbf{96.4}\\
\hline
Bottle & \texttt{concat} & \checkmark & \checkmark & \checkmark & \checkmark & 7.6 & 6.0 & 86.5
 & 94.8 & 96.4 & 93.1 \\
Bottle & \texttt{ours} & \checkmark & \checkmark & \checkmark & \checkmark & \textbf{5.4} & \textbf{4.5} & 
\textbf{92.1} & \textbf{99.0} & \textbf{99.6} & \textbf{97.5}\\
\shline
\end{tabular}
}
\label{tab:fusion_type}
\end{table}

\subsection{Other Ablations}
\noindent\textbf{Ablation on Detector. }
We train an object detector using Faster R-CNN without additional modification of polarimetric inputs. It is not affected by the photometric challenges of the objects, as indicated by similar results in Tab.~\ref{tab:bbox} when training/testing PPP-Net with the GT bounding box and the predicted ones.
\begin{table}[!h]
\caption{\textbf{BBox Ablations. }}
\centering

\resizebox{0.85\columnwidth}{!}{
\begin{tabular}{l|c|c|c|c|c|c}
\shline
\multirow{1}{*}{Configuration} & \multirow{1}{*}{{Cup}}& \multirow{1}{*}{{Teapot}} & \multirow{1}{*}{{Can}} & \multirow{1}{*}{Fork} & \multirow{1}{*}{{Knife}} & \multirow{1}{*}{{Bottle}} \\
\hline
Train with GT BBox/Test with pred BBox & 97.2 & 99.9 & 98.4 & 95.9 & 96.4 & 97.5\\
\hline
Train/Test with GT BBox & 99.0 & 99.9 & 99.0 & 96.1 & 97.6 & 97.5 \\
\shline
\end{tabular}
}
\label{tab:bbox}
\end{table}

\noindent\textbf{Ablation on Refractive Index. }
As mentioned, the prior knowledge of the refractive index of materials in the scene is one limitation of our model. To analyse the impact of incorrect indices, we report pose accuracy results when trained/tested with minor (1.54 vs. 1.5) and large deviations (2.75 vs. 1.5) of the correct index in Tab.~\ref{tab:index}. The results in the 2nd row highlight that our model still performs well when providing incorrect refractive indices during inference. This indicates that the model is robust enough to extract relevant features. When training and testing with very different indices, we see a slight decrease in ADD (cf. \textit{fork, knife}).
\begin{table}[!h]
\caption{\textbf{Refractive Index Ablation. }}
\centering

\resizebox{0.9\columnwidth}{!}{
\begin{tabular}{l|c|c|c|c|c|c}
\shline
\multirow{1}{*}{Object} & \multirow{1}{*}{{Cup}} &\multirow{1}{*}{{Teapot}} & \multirow{1}{*}{{Can}} & \multirow{1}{*}{Fork}& \multirow{1}{*}{{Knife}} & \multirow{1}{*}{{Bottle}}\\
Refractive Index & 1.50 & 1.54 & 1.35 & 2.75 & 2.75 & 1.52\\
\shline
Train/Test with correct index & 97.2 & 99.9 & 98.4 & 95.9 & 96.4 & 97.5\\
\hline
\multirow{2}{*}{\shortstack{Train with correct index, \\test with incorrect (1.5)}} & \multirow{2}{*}{97.2} & \multirow{2}{*}{99.9} & \multirow{2}{*}{98.3} & \multirow{2}{*}{95.8} & \multirow{2}{*}{96.2} & \multirow{2}{*}{97.5}\\&&&&&\\
\hline
Train/Test with incorrect index (1.5) & 97.2 & 99.9 & 98.0 & 93.5& 90.1 & 97.5\\
\shline
\end{tabular}
}
\label{tab:index}
\end{table}

\noindent\textbf{Ablation on Photometric Complexity. }
Recent RGB-D pipelines (to which we compare) try to overcome photometric challenges, e.g. textureless objects, by incorporating depth information.
Correct depth information is essential here, but depth sensors suffer specifically in these areas.
On the contrary, the physical properties encoded in the polarimetric images, which are leveraged by \textbf{PPP-Net}, preserve object-centric shape information also for very challenging (e.g. reflective, transparent) objects. 
We train and test CosyPose~\cite{labbe2020cosypose} on our data using single-view mode without ICP refinement or additional 1 million synthetic data as when training on T-LESS~\cite{hodan2017t}, which ensures the same settings for all benchmarking experiments. We outperform CosyPose for every object in Tab.~\ref{tab:cosypose_benchmark} with significant improvement for photometrically challenging objects.

\begin{table}[!h]
\caption{\textbf{CosyPose~\cite{labbe2020cosypose} Benchmarking. }}%
\centering

\resizebox{0.65\columnwidth}{!}{
\begin{tabular}{l|c|c|c|c|c|c|c}
\shline
  \small Methods &
  \rotatebox[origin=c]{0}{\small Cup}          &
  \rotatebox[origin=c]{0}{\small Teapot}        &
  \rotatebox[origin=c]{0}{\small Can} &
  \rotatebox[origin=c]{0}{\small Fork}   &
  \rotatebox[origin=c]{0}{\small Knife}    &
  \rotatebox[origin=c]{0}{\small Bottle}     &
  \rotatebox[origin=c]{0}{\small Mean}  

  \\ 
\hline
CosyPose [33]    & 88.5 & 94.3 &     91.0     &        83.0      &  89.5  &   79.6   & 87.7     \\
\hline
Ours         & \textbf{97.2} & \textbf{99.9}   & \textbf{98.4} & \textbf{95.9} & \textbf{96.4} & \textbf{97.5} & \textbf{97.6}    \\
\shline
\end{tabular}
}
\label{tab:cosypose_benchmark}
\end{table}

\section{Qualitative Visualizations}
\begin{figure*}[!ht]
    \begin{center}
    \includegraphics[width=1.0\linewidth]{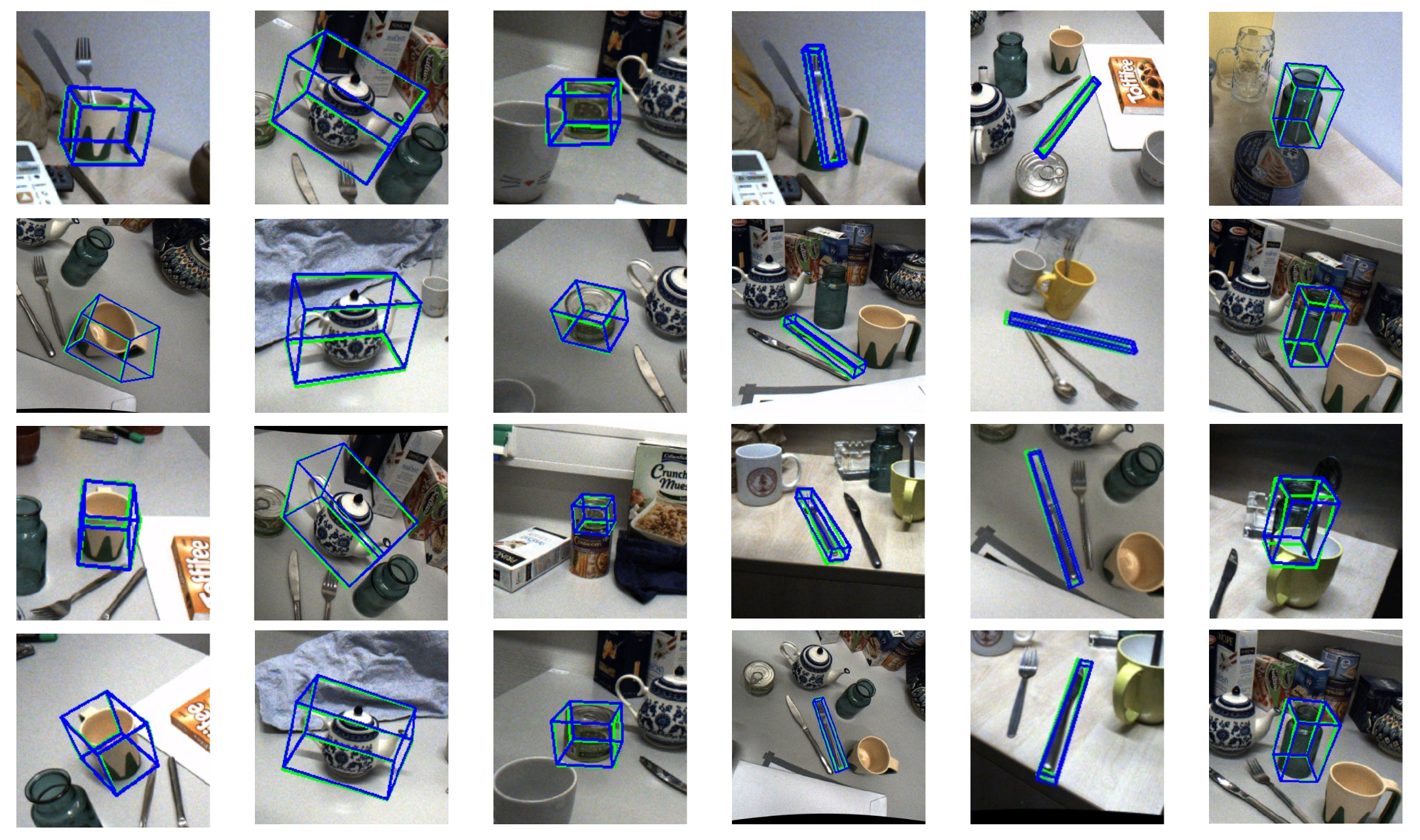}
    \end{center}
     \caption{\textbf{Qualitative Results.} Predicted and GT 6D poses are illustrated by \textit{blue} and \textit{green} bounding boxes, respectively.
      }
    \label{fig:supp_qual}
\end{figure*}

In Fig.~\ref{fig:supp_qual}, we visualise the 6D pose by overlaying the image with the corresponding transformed 3D bounding box. For better visualization we cropped the images and zoomed into the area of interest.

\section{Instance-level Polarimetric Object Pose Dataset}
\begin{figure*}[!ht]
    \begin{center}
    \includegraphics[width=\linewidth]{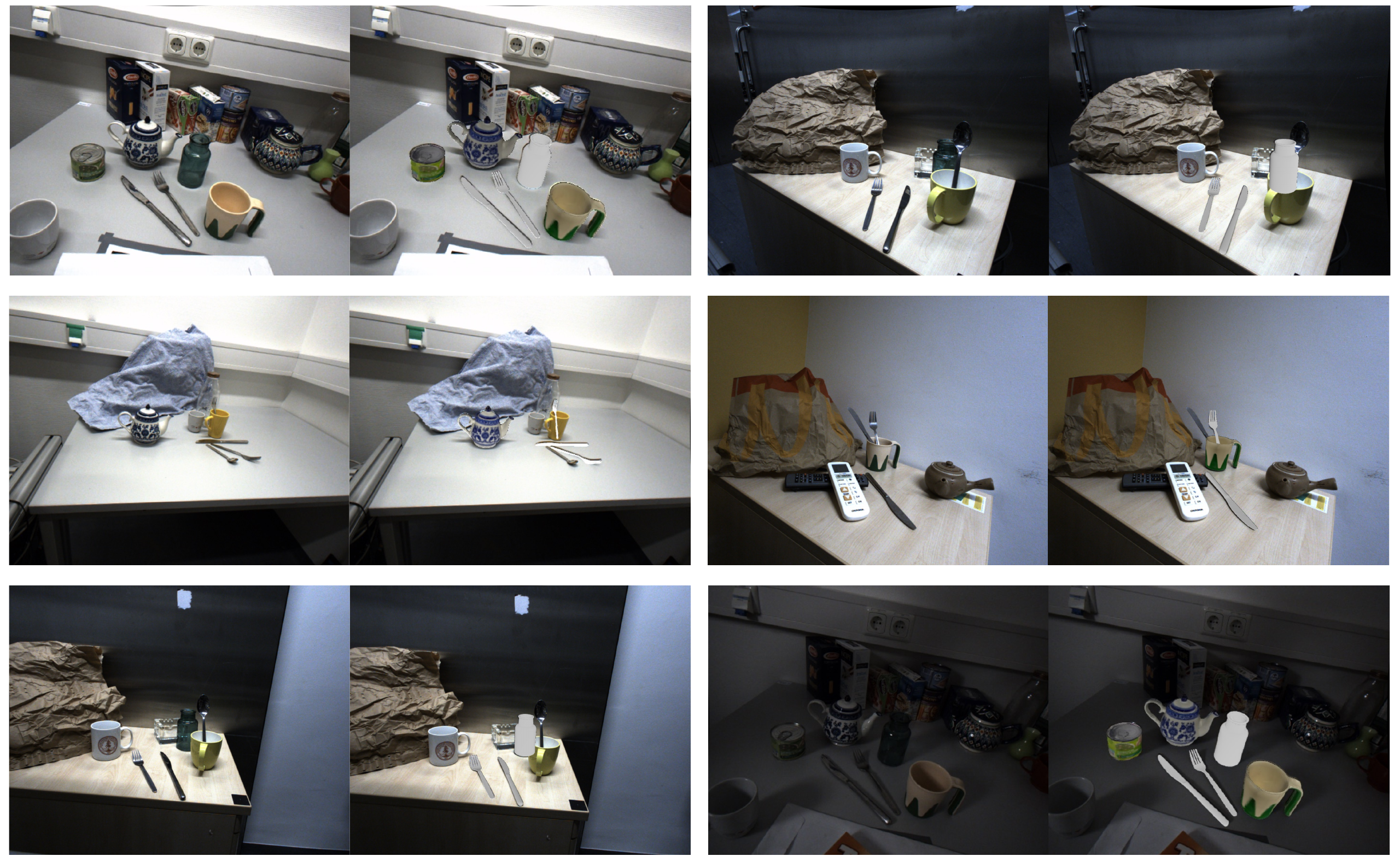}
    \end{center}
     \caption{\textbf{Dataset and Annotation Examples.} The figure shows one polarisation image together with the rendered models.
      }
    \label{fig:dataset}
\end{figure*}

Fig.~\ref{fig:dataset} illustrates our scene settings as well as the pose annotation quality. We cover a wide range of variety in the background, illumination, as well as object settings. And our pose annotations are accurate for all objects, including the challenging reflective and transparent ones. High accuracy of annotations is achieved with the process described in~\cite{PhoCal}, which involves tipping multiple times the surface of objects with a calibrated tool tip attached to a robotic arm and subsequent ICP alignment with the pre-scanned 3D mesh of the object (see Sec.4 for more details). We provide 6D pose annotations for all objects in the scene, but here only consider the objects introduced in Fig. 5 which cover a wide range of photometric complexity. Fig.~\ref{fig:dataset} shows the superimposed 3D meshes of these objects with high accuracy.

\noindent\textbf{Camera Alignment.} The extrinsic calibration, which is derived by an hand-eye calibration against the robotic end-effector with high accuracy, is used for aligning different camera modalities. The alignment of cameras is only limited by their form factors. To reduce this effect and to bring the optical centers of all cameras as close to another as possible, we design a custom rig. However, small changes in the viewpoint of each camera cannot be completely avoided.

\noindent\textbf{Dataset Comparison.} Tab.~\ref{tab:dataset_comparison} gives an overview of different dataset characteristics. 
\begin{table}[!hb]
\caption{\textbf{Dataset Comparison. }}%
\centering

\resizebox{1.0\columnwidth}{!}{
\begin{tabular}{l|ccc|c|cccc|c}
\shline
  \small Dataset &
  \rotatebox[origin=c]{45}{\small RGB}          &
  \rotatebox[origin=c]{45}{\small Depth}        &
  \rotatebox[origin=c]{45}{\small Polarisation} &
  \rotatebox[origin=c]{45}{\small Robotic GT}   &
  \rotatebox[origin=c]{45}{\small Occlusion}    &
  \rotatebox[origin=c]{45}{\small Symmetry}     &
  \rotatebox[origin=c]{45}{\small Transparent}  &
  \rotatebox[origin=c]{45}{\small Reflective}   &
  \rotatebox[origin=c]{45}{\small Sequences}      
  \\ 
\hline
YCB-V [55]      & \checkmark & \checkmark &          &              & \checkmark & \checkmark &        &             & $92$      \\
T-LESS [23]      & \checkmark & \checkmark &          &              & \checkmark & \checkmark &        &             & $20$      \\
Linemod [21] & \checkmark & \checkmark &           &              & \checkmark & \checkmark &        &              & $15$     \\
\hline
Ours         & \checkmark & \checkmark   & \checkmark & \checkmark & \checkmark & \checkmark & \checkmark & \checkmark     & $20$    \\
\shline
\end{tabular}
}
\label{tab:dataset_comparison}%
\end{table}

\clearpage

\end{document}